\documentclass[11pt]{article}
\usepackage[margin=1in]{geometry}
\usepackage[T1]{fontenc}
\usepackage[utf8]{inputenc}
\usepackage{amsmath,amssymb}
\usepackage{graphicx}
\usepackage{booktabs}
\usepackage{array}
\usepackage{microtype}
\usepackage[numbers,sort&compress]{natbib}
\usepackage{xcolor}
\usepackage{colortbl}
\usepackage{adjustbox}
\usepackage[colorlinks=true,linkcolor=blue,citecolor=blue,urlcolor=blue]{hyperref}
\newcolumntype{P}[1]{>{\raggedright\arraybackslash}p{#1}}
\newcolumntype{I}[1]{>{\hspace{1.1em}\raggedright\arraybackslash}p{#1}}
\definecolor{owmiband}{gray}{0.90}
\definecolor{owmirow}{gray}{0.965}

\title{Open-Weight Masked Introspection:\\ Measuring What Language Models Can Report\\ About Their Own Computation}
\author{Emilio Ferrara\\ University of Southern California\\ \texttt{emiliofe@usc.edu}}
\date{\today}

\begin{document}
\maketitle

\begin{abstract}
Are frontier models able to introspect about their internal states? Recent work suggests that under certain conditions a complex enough model can audit its own internals, call out what changed, and report back confidently about it. We tested that claim on eight open-weight models from seven families and found no such ability: asked whether their own computation had been altered, none answered better than chance. To test it we built Open-Weight Masked Introspection (OWMI), a framework that intervenes on internal computational objects such as residual-stream sites, attention heads and sparse-autoencoder features, then interrogates the model about the change against the null conditions an answer has to beat: sham runs where nothing was altered, impact-matched random perturbations, and a text-only observer that sees only the visible output. OWMI attaches to benchmarks already in use.

Over 78{,}000 measurements, no model's report discriminates a real intervention from a sham beyond chance (AUROC $\approx 0.5007$), and an equivalence test bounds the effect below $0.15$ percentage points of AUROC. Surprisingly, all the information needed to make that assessment is in the models. A model fine-tuned to report this class of intervention reaches near-perfect recovery on held-out directions (AUROC $\approx 1.0$), and a linear probe recovers intervention presence from the same activations at between 75\% and 95.8\% accuracy, sharpening to no held-out error at the last layer before the model speaks. In one model the signal surfaces in the confidence rather than the words: its yes-or-no report never varies, while the confidence attached to that report separates intervention from sham at AUROC $0.647$. The failure sits in the path from internal state to verbal report, so oversight practices that read a model's own testimony, from chain-of-thought monitoring to self-critique, need validating against an internal reference rather than against that testimony.

While our results show the inability of current open-weight models to introspect, the debate is not settled for future models. We release OWMI as a library so that this emerging ability can be measured as it develops: \url{https://huggingface.co/emilioferrara/owmi}.
\end{abstract}

\section{Introduction}\label{sec:intro}

Can a language model report what happened inside the computation that produced its answer?

A growing set of oversight practices assumes it can. Chain-of-thought monitoring assumes that stated reasoning tracks performed reasoning. Self-critique assumes that a model can audit the process behind its output. Confidence elicitation assumes that a model can read its own uncertainty \citep{turpin2023unfaithful,lanham2023faithfulness,kadavath2022mostly}. Recent intervention studies encourage the assumption: inject a concept into a model's activations and it will sometimes notice that something is wrong, and occasionally name it \citep{lindsey2025introspection}. We tested that capability systematically on eight open-weight models from seven laboratory families and reached a confident conclusion: no model reports our controlled interventions at a magnitude that supports detection.

Nothing in the benchmark ecosystem would have caught this, and the reason is structural. Benchmarks grade the answer a model gives. None of them grades what the model can say about the computation that produced the answer, because none of them knows what that computation was. Without a reference, a report that is true and a report that merely sounds true score identically, and a model that always claims to notice something does as well as a model that genuinely does. The intervention studies supply that reference and establish the sham condition that goes with it; what they leave open is how much of a report survives once generic disruption and plain output reading are also ruled out.

\begin{figure}[t]
  \centering
  \includegraphics[width=\linewidth]{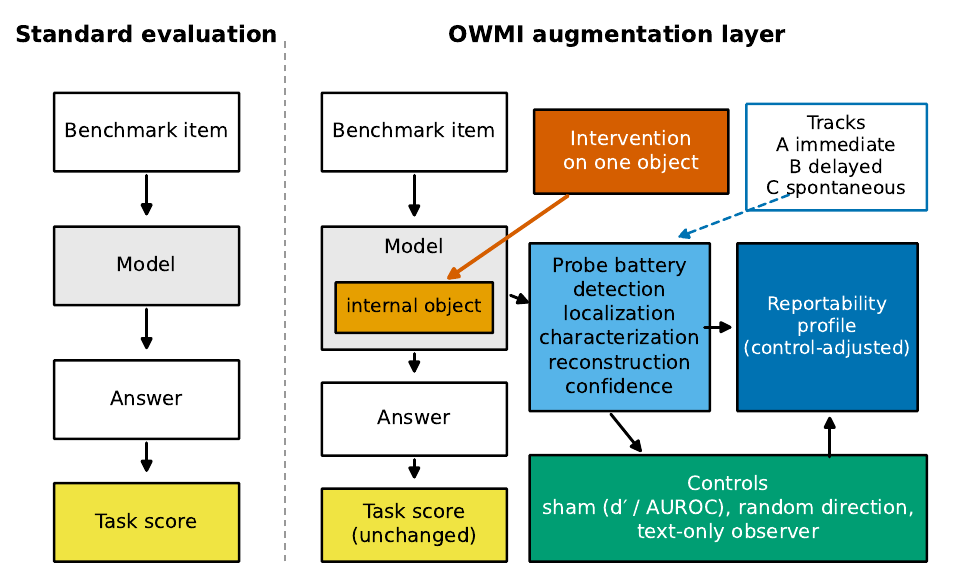}
  \caption{The OWMI measurement loop. A standard benchmark item (left) is evaluated normally, producing a task score. OWMI (right) wraps the same item: a baseline forward pass is cached, a controlled intervention is applied to one computational object (residual site, attention head, or SAE feature), and an introspective probe battery (detection, localization, characterization, reconstruction, confidence) is administered under one of three temporal tracks. Sham runs, matched random perturbations, and a text-only observer bound the scoring. The benchmark's own task scoring is preserved unchanged.}
  \label{fig:hero}
\end{figure}

We built the measurement around those two questions. Open-Weight Masked Introspection (OWMI, pronounced ``owe me'') takes an ordinary benchmark item, preserves its task and its scoring, caches a baseline forward pass, alters one internal computational object, and then asks the model what changed (Figure~\ref{fig:hero}). The altered object is a residual-stream site, an attention head, or a sparse-autoencoder feature, so the ground truth is imposed rather than inferred: we know what was done, where, and how hard. Three null conditions set what an answer has to beat. Sham runs alter nothing, so a model that always reports a change scores at chance. Random perturbations matched for downstream disruption withhold credit for noticing generic damage rather than the specific object. A text-only copy of the model, shown only the visible output, marks the level a report must exceed to carry more than an outside reader could already infer. OWMI supplies the interventions, the probes and the scoring, not the prompts, so it adds intervention-reportability measurements to benchmarks already in use without replacing their prompts or task scores.

We call the measured property intervention reportability: the degree to which a model's output carries information about a controlled change to its own computation. This is a claim about information flow, not about consciousness or self-awareness. The distinction is what the controls enforce. A model may react to a perturbation because the perturbation changed its output, or describe something unusual because the prompt invited that answer, without reading the internal event at all. Sham discrimination, impact matching and the observer bound remove those explanations in stages, leaving a narrow and testable question: when we alter a known internal object, does the model's report carry information about that alteration which the prompt, the visible output, and generic disruption cannot explain?

The answer is no. Our experiments comprise over 78,000 measurements on models from 0.5B to 15B parameters. Pooled over the 11,216 paired trials of the dose battery, report discrimination reaches an AUROC of $\approx 0.5007$, an advantage of seven hundredths of a percentage point, and equivalent to guessing: the discrimination advantage is bounded below $0.15$ percentage points of AUROC ($p < 0.0001$). A separate breadth battery adds 19,520 measurements on three models, crossing three classes of computational object, six depths from layer 8 to layer 31, four benchmark domains, both executed temporal tracks and four scored probe families. No level of any axis changes the conclusion.

A null result invites two easy dismissals, and we close both. The first is that the instrument is too blunt to register anything. It is not: a model fine-tuned to make exactly this report clears the identical pipeline at $d' = 5.15$ and AUROC $\approx 1.0$, identifying all 100 intervention trials and all 100 sham trials on directions absent from its training. The second is that there is nothing to report. There is: a linear probe reading the same activations recovers intervention presence with held-out accuracy of 95.8\% and 75.0\% in the two dose-calibrated models, against a chance level of 50\% and a label-shuffled null that never reaches those values in 200 refits. Re-harvesting the same probe downstream of the intervention, at every deeper layer we tested and up to the last layer before the model writes its answer, separates intervention from sham with no held-out error, so the signal is not merely present where we perturbed but still fully available where a report would have to be formed.

This is the result we want carried forward. The information is in the activations, and the model's own account of itself does not carry it. The failure sits in the path from internal state to verbal report, not in the availability of the information and not in our ability to measure it. That distinction decides what would fix the problem: an oversight scheme that reads activations works with a signal that is present, while one that asks the model has so far obtained none of it.

One channel is not silent, and it makes the picture more interesting than a flat null. In Qwen2.5-7B-Instruct the discrete answer to ``did anything change'' is constant across every scorable trial, so that report discriminates at an AUROC of exactly 0.500, while the confidence attached to it discriminates intervention from sham at 0.647: the model is markedly less confident when it has been intervened upon. The signal reaches a graded quantity the model emits without reaching the words it chooses. Our null therefore concerns the discrete report, which is what an oversight scheme reads when it asks a model what happened.

Our results establish the inability of current open-weight models to introspect on this class of internal event, at the sites and scales we measured. Reconstruction has not yet been scored and the spontaneous track produced no complete intervention-sham pair, so neither supports an estimate here, and the question stays open for larger models, closed-weight systems, and architectures trained to build the missing reporting path. That is why we release the instrument and not only the finding. We release OWMI so that any claim of introspective access, present or future, is scored against an internal reference rather than against the model's own testimony.

\section{Related Work}\label{sec:related}

\textbf{Introspection and self-report in language models.} The claim we test has a specific origin. \citet{lindsey2025introspection} injected a concept vector into a large model's activations and found that the model can sometimes report that something unusual occurred, and can occasionally name the concept that was injected, against a sham condition that produced no false positives over one hundred trials in production models. We build OWMI's atomic measurement on that concept-injection protocol. Within months, at least five groups took up the same question, and they cite \citet{lindsey2025introspection} rather than each other, which makes the cluster a set of independent investigations rather than a single line of commentary. Models detect \emph{that} an intervention happened far more reliably than they identify \emph{what} it was, and the binary detection paradigm is confounded by an intervention-induced shift toward affirmative answers \citep{hahami2025disturbance}. Apparent introspection may reduce to first-order anomaly detection rather than second-order access, since models cannot reliably separate an intervention on their internal states from a manipulation of their input \citep{singh2026realitycheck}. The effect appears to be content-agnostic \citep{lederman2026contentagnostic}, it has been traced mechanistically \citep{macar2026mechanisms}, it survives a protocol that removes the perturbation before the report is elicited \citep{pearsonvogel2026latent}, and it has been measured with graded numeric self-reports under a dose ladder \citep{martorell2026quantitative}. Models can also be explicitly trained to report activation steering on held-out concepts \citep{fonseca2025steeringawareness} or to report implanted behaviors \citep{shenoy2026introspectionadapters}. A parallel line approaches introspection through self-prediction, asking whether models can predict their own behavior better than observers with the same data \citep{binder2024lookinginward,betley2025tellme,naphade2026memyself}, and \citet{song2025privileged} makes the observer explicit by asking whether a second model reading the same output does as well.

Sham conditions are therefore established practice in this literature, and we do not claim otherwise. What we have not found assembled in one measurement is the full set of controls. Random controls in this literature are matched on the norm of the injected vector \citep{lindsey2025introspection,fonseca2025steeringawareness,martorell2026quantitative} or on the semantic content of the manipulation \citep{singh2026realitycheck}, and a norm-matched control equates how large the perturbation was rather than how much damage it did. Observer baselines, where they exist, read the model's \emph{input} \citep{singh2026realitycheck} or are supplied by the same steered model in a third-person framing \citep{lederman2026contentagnostic}; the one genuine independent observer of the visible output studies a sampling hyperparameter rather than an intervened internal object \citep{song2025privileged}. We are not aware of a design that scores a report against a sham, an impact-matched random perturbation, and a text-only observer restricted to the visible output, alongside a fine-tuned known positive and a linear probe on the same activations. OWMI specifies that combination, and we are explicit about where our own batteries realize it and where they do not: the impact-matched control is realized for one model, approached but overshot for a second, and unit-norm for the other six, and the observer quantities we report come from the weakest member of the observer family (Sections~\ref{sec:controls} and~\ref{sec:res-detection}). The contribution we defend without qualification is narrower and does not depend on those details: an equivalence bound in place of a non-significant difference, which converts a failure to detect into a statement about how large the effect can be. Two of these lines become working parts of our design rather than background. We turn the leakage-versus-access critique of \citet{singh2026realitycheck} into an estimand, and trained steering awareness \citep{fonseca2025steeringawareness}, which reaches high detection accuracy on held-out concepts with no false positives on clean controls, supplies the known positive that validates instrument sensitivity, clearing our pipeline at $d' = 5.15$ on held-out directions.

\textbf{Metacognition measurement in psychophysics.} The problem of asking a system to report on its own processing is not new, and psychophysics has spent four decades on it, separating first-order task performance from second-order discrimination of whether that performance was correct. Type-2 signal detection theory formalizes the distinction, meta-$d'$ is the established estimator of metacognitive sensitivity, and hierarchical estimation extends it to comparisons of metacognitive efficiency across participants and conditions \citep{maniscalco2012signal,fleming2014how,fleming2017hmetad}. The monitoring and control framework of metamemory gives an earlier account of how object-level and metacognitive processes interact \citep{nelson1990metamemory}. The same tradition documents the failure mode our null hypothesis anticipates. Human verbal reports can supply plausible but confabulated accounts of mental processes, and people can fail to notice that the outcome of their own choice was exchanged \citep{nisbett1977telling,johansson2005failure}. No-report paradigms, which separate a target process from reports explicitly elicited during that process, are the direct methodological precedent for Track C \citep{tsuchiya2015noreport}, and higher-order theories provide the conceptual home for second-order access \citep{lau2011empirical}. We adopt this measurement tradition, including its catch trials, from which sham discrimination and the false-alarm rates that signal detection theory is built on are directly inherited rather than invented here. What the human paradigms cannot supply is an intervention on the substrate with known ground truth. A human experimenter cannot set a specific internal representation and ask the participant about it, so the perturbation is delivered through the stimulus. Working on open weights lets us impose the internal event itself, which is what makes an impact-matched perturbation of that event definable at all.

\textbf{Interventional interpretability.} We use established intervention methods: causal mediation analysis \citep{vig2020causal}; localization and editing through activation patching \citep{meng2022rome}, applied under established methodological guidance \citep{zhang2024activationpatching}; and sparse autoencoders as dictionaries of interpretable feature directions \citep{cunningham2024sae,bricken2023monosemanticity,templeton2024scaling}, with public annotation tools \citep{lin2023neuronpedia}. Activation-space directions can alter model behavior without parameter updates, as demonstrated across representation engineering, activation engineering, contrastive activation addition, inference-time intervention, and latent steering vectors \citep{zou2023representation,turner2023activation,rimsky2024steering,li2023inference,subramani2022extracting}. Calibrated residual-stream injections extend this paradigm to fine-grained control over psychological traits with reliable, near-linear dose-response behavior \citep{blas2026psychological}, a precedent for the dose ladder we use to calibrate intervention strength (Section~\ref{sec:interventions}). One finding from this literature bears directly on our controls. The Hydra effect occurs when a model compensates elsewhere in its computation after a component is ablated \citep{mcgrath2023hydra}. This self-repair matters for impact matching and for our causal-relevance check because a locally strong intervention can produce a weak output effect even when it targets an active object. We depart from this literature in what the intervention is for. Interpretability treats an intervention as a probe the experimenter reads. We treat the same intervention as a stimulus and ask what the model can read.

\textbf{Metacognition and faithfulness of self-report.} Evidence about what a model knows of its own behavior points two ways at once, and the tension is the reason our controls exist. On one side, models can partially evaluate whether their own answers are correct \citep{kadavath2022mostly} and can express confidence with varying degrees of calibration \citep{tian2023justask,xiong2024uncertainty}; verbalized uncertainty can be explicitly taught \citep{lin2022teaching}, and semantic uncertainty and semantic entropy estimate uncertainty over meanings rather than surface forms \citep{kuhn2023semantic,farquhar2024detecting}, with \citet{geng2024survey} reviewing the broader confidence-estimation and calibration literature. Truthfulness-related information can likewise be recovered from internal states with supervised or unsupervised probes \citep{azaria2023internal,burns2023discovering}, a precedent for our linear-probe anchor, which asks whether intervention information exists in a readily extractable internal form before we interpret verbal access. SAD and mathematical metacognition evaluations add complementary benchmarks of situational self-knowledge and self-evaluation \citep{laine2024sad,didolkar2024metacognitive}. By contrast, chain-of-thought explanations systematically misreport the causes of model behavior \citep{turpin2023unfaithful,lanham2023faithfulness}. The reverse dissociation also occurs: a reasoning chain can contain information that the final output systematically omits, as documented for politically sensitive content in DeepSeek-R1 \citep{qiu2025suppression}. Internal traces can contain what outputs omit, and outputs can claim what internal traces do not support. Current models therefore combine calibrated self-evaluation with unfaithful self-explanation. Our controls test that tension against the computation itself.

\textbf{Evaluation infrastructure.} We write no benchmark prompts of our own. OWMI attaches to the existing evaluation ecosystem, principally lm-evaluation-harness \citep{gao2024lmevalharness} and OpenCompass \citep{opencompass2023}, and leaves each source benchmark's items and scoring untouched.

\section{Conceptual Framework}\label{sec:framework}

\subsection{The atomic measurement}

An OWMI item is defined by
\begin{equation}
  (p_i,\; o_j,\; a_k,\; q_\ell,\; s_\ell),
\end{equation}
where $p_i$ is a source benchmark prompt, $o_j$ is a computational object such as a residual-stream site or an attention head, $a_k$ is an intervention operator that alters it, $q_\ell$ is an introspective probe put to the model, and $s_\ell$ is the function that scores the answer. Each realized measurement also records the layer, temporal track, and seed used for aggregation. We run each item once without an intervention and once with it. The first run gives us the baseline trace and external answer; the second gives us the altered trace and answer. We evaluate the probe on the altered run and compare it with the baseline and matched controls.

The framework distinguishes three related outcomes:
\begin{itemize}
  \item \textbf{Task performance:} whether the original benchmark answer remains correct.
  \item \textbf{Intervention sensitivity:} whether the intervention changes the model's behavior or representation.
  \item \textbf{Intervention reportability:} whether the model can report properties of that change under controls that prevent the answer from being inferred from the prompt alone.
\end{itemize}
These outcomes need not travel together. A model may react to a perturbation without describing it, or give a plausible introspective account that does not track the altered computation. In the latter case, the probe wording may reveal the intervention, or the output may contain a recognizable effect of it.

\subsection{Introspection battery}

For each intervention we administer a matched battery of five questions, which ask in turn whether the computation changed (\textbf{detection}), how sure the model is of that answer (\textbf{confidence}), where the change occurred (\textbf{localization}), what kind of thing changed (\textbf{characterization}), and what the change took away (\textbf{reconstruction}).

One number cannot represent all five capabilities, so we score each probe family on its own terms. Detection and localization use discrimination and calibrated classification accuracy against explicit chance levels. Characterization uses category accuracy and macro-F1 over a closed set matched to the benchmark domains. Reconstruction combines exact match, category match, and embedding similarity. Similarity metrics reward topical paraphrase, which is credit for sounding right, so we add human validation on a subset. We evaluate confidence through calibration and selective prediction \citep{tian2023justask,xiong2024uncertainty}, never through mean self-reported confidence.

\subsection{Three temporal tracks}

\textbf{Track A: Immediate.} The introspective query follows the intervention directly, before the model has generated any task output. Nothing visible exists yet for the model to read, so this track estimates the upper bound on reportability under an explicit cue.

\textbf{Track B: Delayed.} The model first completes the primary task and then receives a reflection prompt whose context contains the task prompt and the model's own task output. The baseline side of each pair reflects on its own baseline output, so the false-alarm reference remains a true no-intervention context. This track tests whether information about the altered computation survives ordinary generation and is still reportable afterward. A masked-context variant strips the task prompt and output from the reflection context, which separates delay itself from the model reading its own text.

\textbf{Track C: Spontaneous.} A third track asks the model to reflect with no intervention cue and no introspective question, so that an anomaly report must be volunteered rather than prompted. It is the strongest form of the capability. The track ran but does not support the paired estimator used here, so we describe it and its data in Appendix~\ref{app:excluded} and leave its analysis to future work.

\subsection{Controls and adjusted scoring}\label{sec:controls}

Three controls set what an answer has to beat. Each rules out one alternative explanation, and each changes the score accordingly.

\textbf{Sham runs.} In every (model, benchmark, object, operator) cell, a matched fraction of runs receives the full probe battery with no intervention, paired on the same items. This control separates introspective sensitivity from acquiescence, the policy of answering yes on every run whatever happened. We therefore score discrimination between paired intervention and sham responses, never the raw yes-rate. With hit rate $H$ on intervention runs and false-alarm rate $F$ on paired shams,
\begin{equation}\label{eq:dprime}
  d' = \Phi^{-1}(H) - \Phi^{-1}(F),
  \qquad
  \mathrm{AUROC} = \Pr\!\left(g_{\mathrm{int}} > g_{\mathrm{sham}}\right) + \tfrac{1}{2}\Pr\!\left(g_{\mathrm{int}} = g_{\mathrm{sham}}\right),
\end{equation}
where $\Phi^{-1}$ is the probit and $g$ is a graded detection score; Appendix~\ref{app:formal} gives the estimation details. A model that always reports ``something changed'' attains $H = F = 1$ and $d' = 0$. Its raw detection is perfect, but its discrimination is exactly at chance. Sham runs also supply the false-alarm base rates that Track C requires, because a spontaneous anomaly report is evidence only relative to how often the model volunteers anomalies when no intervention occurred.

\textbf{Impact-matched random perturbations.} For every directional intervention, we apply a random direction at the same site and calibrate it to match the treatment's downstream effect, not merely its norm. Norm matching alone systematically understates the control. A random direction in a high-dimensional residual space lands nearly orthogonal to the feature manifold, so it does less damage at the same norm and makes the treatment look more distinctive than it is. Matching on effect is also demanded by how disproportionately language-model behavior can respond to nominally small perturbations, where semantically equivalent input changes cascade into large behavioral shifts \citep{ye2026butterfly}. This control separates noticing the particular object we altered from noticing that the computation was disturbed at all. The distinction is the reason we depart from established practice here: random controls in this literature are matched on the norm of the injected vector \citep{lindsey2025introspection,fonseca2025steeringawareness,martorell2026quantitative}, which equates how large the perturbation was rather than how much it disturbed the computation, and it is the second quantity that a claim of specific access has to survive. We score each probe family as a margin over the matched perturbation. Non-directional operators receive matched controls with the same dimensionality and calibrated impact.

\textbf{Text-only observer bound.} We give a separate, unintervened model instance the visible record of the intervened run and a matched probe battery. The observer holds everything the visible text contains and nothing private to the intervened model. The control rules out output reading, which matters most in Tracks B and C, where the intervened model reads its own possibly degraded output before the probe arrives. Whatever the text alone yields is observation, not introspection. Genuine reportability must therefore clear the observer bound, and clearing chance is not enough. The bound is defined over the visible \emph{output}, which is what distinguishes it from the input-side baselines used elsewhere. A classifier reading the model's input bounds what the prompt already predicts \citep{singh2026realitycheck}; a third-person reframing run on the same steered model is not independent of it \citep{lederman2026contentagnostic}. Reading the output is the case that matters for oversight, because an overseer reads outputs. \citet{song2025privileged} construct such an independent reader, but over a sampling hyperparameter rather than an intervened internal object. We define the bound as the maximum performance of three increasingly strong observers: the same model zero-shot, a stronger open-weight model, and a supervised classifier trained on held-out visible outputs. The batteries reported here execute the first of the three, so every observer quantity in Section~\ref{sec:results} is the same-model zero-shot observer rather than the full maximum. That is the weakest member of the family, so the bound we report is a lower bound on the observer bound, and a report that failed to clear it would fail to clear the stronger observers as well. Since no model clears even this observer, the direction of the conclusion is unaffected; a positive result would have required the full maximum before it could be believed. Because Track A has no task output at probe time, its observer receives only the task prompt. That makes the Track A observer an input-side baseline of the kind used elsewhere in this literature rather than the output-side control described here, and we do not count it toward the latter. The distinction between bounding what the prompt predicts and bounding what a reader of the answer could infer applies only to Tracks B and C, where a visible output exists at probe time. We reword observer probes from first person to third person under a fixed matching rule and release both versions side by side for audit.

For each probe family, we report three answers: whether the report beats sham, an impact-matched perturbation, and the observer bound. We never collapse this profile to a single number. A composite index would weight the probe families against each other, and nothing in the data tells us what those weights should be.

\subsection{Testing for equivalence rather than for a difference}\label{sec:equivalence}

A null result cannot be established by failing to reject a nil hypothesis, and at campaign sample sizes
that failure does not occur in any case. With more than eleven thousand paired trials, an effect far too
small to matter will still produce an interval that excludes chance, so a test of $H_0\!:\!\mathrm{AUROC}
= 0.5$ answers a question we are not asking. We test equivalence instead, with two one-sided tests
against an interval of practical indifference \citep{lakens2017equivalence}: for a margin $\delta$, the
procedure tests $H_0^{+}\!:\!\mathrm{AUROC} \geq 0.5 + \delta$ and $H_0^{-}\!:\!\mathrm{AUROC} \leq 0.5 -
\delta$ and takes the larger of the two one-sided $p$-values, so rejecting both places the effect inside
the indifference interval.

Rather than fix $\delta$ and report whether the test clears it, we inverted the procedure and report the
tightest margin the data support. For a chosen level $\alpha$, the smallest $\delta$ at which both
one-sided tests reject is $\hat{\theta} - 0.5 + z_{1-\alpha}\,\mathrm{SE}$, which is the quantity we
quote, expressed in percentage points of AUROC. This reports how tightly the effect is bounded rather than whether it clears a threshold we
selected, and it removes the choice of $\delta$ from our hands. Standard errors come from the same
item-pair clustered bootstrap used for every interval in this paper, so the equivalence bound and the
interval estimates rest on one resampling scheme.

\subsection{First-order leakage and second-order access}\label{sec:leakage}

Even a strong OWMI score would leave one question open: did the model report the change, or did the change simply push the report toward certain tokens? When an intervened forward pass produces a report, the intervention lies in the causal chain that generated the report's tokens, and a perturbed hidden state can bias the output distribution directly. The model may therefore emit ``something felt unusual about arithmetic'' for the same reason that it emits degraded arithmetic: the perturbation propagated to the logits. We call this direct propagation from the perturbed state to the report \textbf{first-order leakage}. An introspection claim needs the stronger condition we call \textbf{second-order access}, in which one part of the computation reads, summarizes, or otherwise uses the state of another, so that the report represents the change instead of merely resulting from it.

We use a minimal definition of second-order access. A learned associative pathway counts if it satisfies this informational test; richer interpretations require evidence beyond this design.

A third mechanism lies between leakage and access. Because the probe question shares a context with the intervened computation, an active intervention can alter how the model encodes the question itself. The model may then answer a subtly different question rather than report on its state. We call this \textbf{probe-encoding perturbation}. It is neither leakage nor access: nothing travels through the task output, and nothing is read. The lifted-at-probe temporal scope removes it by construction, since the model processes the probe prompt without intervention. The contrast between scope levels therefore bounds probe-encoding perturbation just as the observer bound limits output reading, and detection that survives under the lifted scope cannot be attributed to it.

The Track A data introduced in Section~\ref{sec:res-detection} give a first read of this contrast for the two dose-calibrated models. Splitting the complete pairs of Qwen2.5-7B-Instruct and Mistral-7B-Instruct-v0.3 by hook temporal scope leaves 12 and 10 complete pairs in each of the active-through-probe and lifted-at-probe scopes. Both models score at chance in both scopes ($d' = 0.0$, AUROC $= 0.500$ throughout). The lifted-at-probe scope shows no residual detection after the design removes probe-encoding perturbation. The active-through-probe scope, which permits that mechanism, also shows no detection. At this site and scale, the contrast leaves probe-encoding perturbation indistinguishable from no detection.

Behavioral measurement cannot fully separate these mechanisms, but a dose-response design can bound their separation. Intervention strength is an experimenter-set, pre-treatment variable. For each site, we run a calibrated dose ladder and record the per-run divergence between baseline and intervened next-token distributions. The scaling operator multiplies the activation by the strength parameter. A strength of one leaves the activation unchanged, and perturbation magnitude grows as strength moves away from one. We therefore order the ladder by perturbation magnitude rather than by the strength value. The estimand is the detection margin as measured output divergence approaches zero. We estimate it with a hierarchical dose-response model instead of conditioning on observed divergence, which is a consequence of treatment. Writing $\bar{D}$ for the per-run mean Jensen--Shannon divergence between baseline and intervened next-token distributions, the estimand and the leakage decomposition are
\begin{equation}\label{eq:m0}
  m_0 = \lim_{\bar{D} \to 0^{+}} \left[ \Pr(y = 1 \mid \mathrm{int}, \bar{D}) - \Pr(y = 1 \mid \mathrm{sham}) \right],
  \qquad
  \lambda = 1 - \frac{m_0}{\bar{m}},
\end{equation}
with $\bar{m}$ the aggregate detection margin and $\lambda$ the first-order-leakage share; the full hierarchical model appears in Appendix~\ref{app:formal}. Detection that survives at doses producing no measurable output shift cannot be explained by the model reading its own degraded output. That is the strongest behavioral evidence for access rather than leakage that this design can produce. A per-run causal-relevance check, defined as the patching effect of the intervened object on the task answer, separates runs in which the object was inert from runs in which it was used but the output was preserved; only the latter support the access inference. Track A strengthens the same logic by querying the model before any task output exists, and the observer bound constrains the leakage explanation from the other side. We therefore read reportability only through these controls. Any claim about a self-model rests on the dissociations, and never on an aggregate detection rate.

\section{OWMI: A Design Space over the Benchmark Ecosystem}\label{sec:augmentation}

OWMI writes no prompts of its own. We take examples from established evaluation suites and leave their answers and scoring logic exactly as they were. The program covers general knowledge, commonsense, science, mathematics, truthfulness, coding, instruction following, and long-context tasks, drawing on MMLU \citep{hendrycks2021mmlu}, MMLU-Pro \citep{wang2024mmlupro}, HellaSwag \citep{zellers2019hellaswag}, WinoGrande \citep{sakaguchi2020winogrande}, ARC-Challenge \citep{clark2018arc}, GPQA \citep{rein2023gpqa}, GSM8K \citep{cobbe2021gsm8k}, MATH \citep{hendrycks2021math}, TruthfulQA \citep{lin2022truthfulqa}, HumanEval \citep{chen2021humaneval}, MBPP \citep{austin2021mbpp}, IFEval \citep{zhou2023ifeval}, and LongBench \citep{bai2024longbench}. The program is designed to cover every benchmark listed above, and we evaluate twelve of them in the experiments of Section~\ref{sec:plan}.

\begin{figure}[t]
  \centering
  \includegraphics[width=\linewidth]{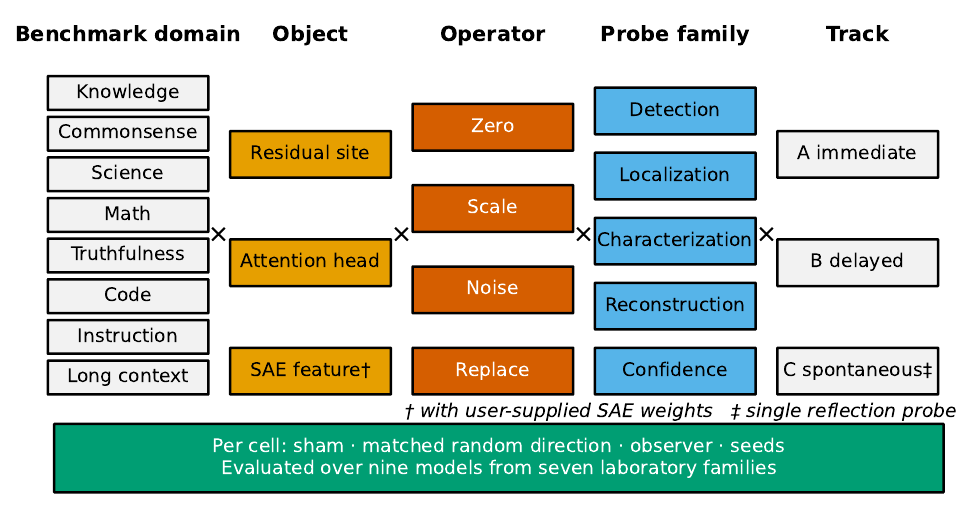}
  \caption{The OWMI measurement space. Each measurement is one cell of the Cartesian product of benchmark domain, computational object, intervention operator, probe family, and temporal track, evaluated per model and seed. Control conditions (sham, matched random direction, text-only observer) attach to every cell.}
  \label{fig:design-space}
\end{figure}

Each source benchmark puts the question to a different computation. MMLU asks for factual retrieval, GSM8K for arithmetic reasoning, HumanEval for program synthesis, and TruthfulQA for the computations implicated in hallucination. What we obtain is therefore a reportability profile, not a single score. The dimensions of this profile carry distinct meanings and support comparison across models, domains, computational objects, intervention operators, and temporal tracks (Figure~\ref{fig:design-space}). The current release makes this comparison across eight models from seven laboratory families. As Section~\ref{sec:intro} explains, domain labels denote measurement contexts rather than established mechanisms. A gap between GSM8K and MMLU could reflect arithmetic against factual computation, or it could reflect nothing more than the surface of the two benchmarks, and we test which.

The augmentation protocol proceeds in seven steps:
\begin{enumerate}
  \item Load a benchmark example and standardize its prompt and answer schema.
  \item Run a baseline forward pass and record the external answer and relevant trace metadata.
  \item Select a computational object and intervention operator.
  \item Run the masked or altered computation.
  \item Administer the introspective battery appropriate to the selected track.
  \item Score task performance, intervention effects, and probe responses separately.
  \item Aggregate by model, benchmark, domain, object, intervention, layer, probe, and seed.
\end{enumerate}

We keep benchmark loading separate from intervention execution. A new source benchmark then costs no change to the introspection procedure, and a new intervention operator applies across every benchmark already supported. Each added benchmark extends the reportability profile while preserving the measurement logic.

We use every benchmark item under its original license, record per-item provenance, and redistribute no benchmark data.

\section{Measurement Protocol and Instrumentation}\label{sec:interventions}

We use forward hooks to alter selected activations while each frozen open-weight model computes. The hooks are temporary functions, so the model weights remain unchanged. We target four classes of computational object: residual-stream sites, attention heads, SAE features, and block outputs. We include an object in the analysis only after object-specific tests validate the intended intervention. Four operators act on these objects: zeroing, scaling, noise injection, and replacement with a reference activation. Each operator's strength parameter is a dose we set and vary, never a constant we tune until something happens.

Every intervention includes an unmodified baseline, a paired sham, an impact-matched perturbation, seeds for stochastic operators, and checks that the selected object exists in the target architecture. We treat hook duration as a design factor with two levels: active through the probe response and lifted when the probe begins. In a decoder that reuses previously computed states, hooks modify initial prompt processing but not later generation steps. In the active-through-probe condition, we intervene again while the model processes the probe prompt. In the lifted-at-probe condition, we leave probe processing unaltered, so any signal must persist from the earlier intervention. The contrast tests whether continued intervention during the probe contributes to the report. Output corruption and malformed generation count as outcomes, never as evidence of introspective access. We run the discrimination analyses both with and without these outcomes and report corruption rates for each cell.

We store every run as a structured record that identifies the model, example, object, intervention, probe, seed, condition, and temporal scope. We run the models on a university Slurm cluster and save the condition, temporal scope, and teacher-forced output divergence. The run specification fixes model revisions, decoding parameters, and probe texts. We never pool quantized and full-precision runs. Quantization changes the very activations our interventions act on, and compression shifts model behavior in ways task accuracy alone does not reveal \citep{ferrara2026quantibias}. Appendix~\ref{app:status} states which parts of the full design the released software implements.

\section{Experimental Evaluation}\label{sec:plan}

We evaluate eight open-weight models drawn from seven laboratory families: Qwen, Mistral, Llama, Gemma, DeepSeek, Phi, and GLM. The evaluation uses twelve benchmarks spanning knowledge, commonsense, arithmetic, code, instruction following, and truthfulness: MMLU, MMLU-Pro, ARC-Challenge, HellaSwag, WinoGrande, GSM8K, MATH, GPQA, HumanEval, MBPP, IFEval, and TruthfulQA. We apply residual-stream interventions in the immediate track under both hook temporal scopes, and for every model the same item runs through all four conditions: intervention, sham, impact-matched random direction, and text-only observer. The design comprises 216 evaluations, one per model, benchmark and hook temporal scope, each contributing 16 measurements from four conditions, two items and two seeds, for 3{,}072 measurement runs and 864 rows in each condition.

For two of the eight models, Qwen2.5-7B-Instruct and Mistral-7B-Instruct-v0.3, we calibrate the random-direction control at each site, use a dose ladder for intervention strength, and sample 32 items per cell. That item count sets the release's execution scope. To learn what it buys, we resample the detection margin in a beta-binomial simulation built on the validation study's item-level variance, with independent item effects on the hit and false-alarm rates. At 32 items, it gives an expected 95 percent confidence interval width of 1.235 $d'$ units at $\alpha = 0.05$ (Appendix~\ref{app:status} reports the full width-by-count curve). The remaining seven models use unit-norm random-direction controls and two seeds in place of this calibrated protocol. Two sensitivity anchors tell us whether a null is informative. On the two dose-calibrated models, a linear probe trained on held-out activations measures how much intervention information is linearly recoverable at each site. A steering-aware fine-tuned model in the style of \citet{fonseca2025steeringawareness} provides the known-positive instrument check. Appendix~\ref{app:analysis} defines the estimands, the exclusion rules, and the scoring for the two dose-calibrated models.

\subsection{Models}

The roster varies laboratory of origin, training provenance, and scale while holding the dense causal decoder family constant, so no difference in reportability across the roster can be attributed to a change of decoder family. The models still differ in depth, hidden width, attention-head configuration, context length and vocabulary (Table~\ref{tab:models}), and those differences are not controlled here. Nine models from seven laboratories place the measurement on independent training pipelines rather than on one family, so a null cannot be an artifact of a single laboratory's post-training. Seven laboratories are a breadth of provenance and not a sample of model space, so the result is a statement about the models we measured rather than about open-weight models in general. We use Qwen2.5-7B-Instruct and Mistral-7B-Instruct-v0.3 for the dose ladder and sensitivity anchors, and Qwen2.5-0.5B-Instruct for the integration check. Table~\ref{tab:models} lists the roster, architecture fields, and LoRA known positive built on Qwen2.5-7B-Instruct. Appendix~\ref{app:status} records hub identifiers, model revisions, gating, and licenses.

\begin{table}[t]
  \centering
  \footnotesize
  \setlength{\tabcolsep}{5pt}
  \caption{Nine evaluated models, ordered by parameter count, and the known-positive LoRA fine-tune of Qwen2.5-7B-Instruct. The fine-tune inherits the base model's architecture unchanged. Params denotes billions of parameters, Hidden the hidden-state width, Q/KV heads the query and key-value head counts under grouped-query attention, Context the configured maximum position count in tokens, and Vocab the tokenizer vocabulary size.}
  \label{tab:models}
  \begin{tabular}{lrrrrrr}
    \toprule
    Model & Params & Layers & Hidden & Q/KV heads & Context & Vocab \\
    \midrule
    Qwen2.5-0.5B-Instruct & 0.49B & 24 & 896 & 14/2 & 32,768 & 151,936 \\
    Mistral-7B-Instruct-v0.3$^{\ddagger}$ & 7.25B & 32 & 4,096 & 32/8 & 32,768 & 32,768 \\
    Qwen2.5-7B-Instruct$^{\ddagger}$ & 7.62B & 28 & 3,584 & 28/4 & 32,768 & 152,064 \\
    Llama-3.1-8B-Instruct$^{\S}$ & 8.03B & 32 & 4,096 & 32/8 & 131,072 & 128,256 \\
    Gemma-2-9B-IT$^{\S}$ & 9.24B & 42 & 3,584 & 16/8 & 8,192$^{\dagger}$ & 256,000 \\
    GLM-4-9B-0414 & 9.40B & 40 & 4,096 & 32/2 & 32,768 & 151,552 \\
    Phi-4 & 14.66B & 40 & 5,120 & 40/10 & 16,384 & 100,352 \\
    DeepSeek-R1-Distill-Qwen-14B & 14.77B & 48 & 5,120 & 40/8 & 131,072 & 152,064 \\
    \midrule
    Known positive (LoRA, Qwen2.5-7B-Instruct)$^{\P}$ & 7.62B & 28 & 3,584 & 28/4 & 32,768 & 152,064 \\
    \bottomrule
  \end{tabular}
  \vspace{2pt}

  \raggedright\footnotesize $^{\ddagger}$Carries the dose-ladder calibration and the linear-probe and known-positive sensitivity checks of Section~\ref{sec:plan}. $^{\S}$Gated repository requiring authenticated access. $^{\dagger}$Gemma-2 alternates sliding-window attention over 4,096 tokens with global attention across layers; the value shown is the configured maximum position count. $^{\P}$Architecture fields are those of the base model; the LoRA fine-tune does not alter them. Its held-out evaluation result appears in Section~\ref{sec:res-sensitivity}, not in this table.
\end{table}

The roster spans MIT, Apache 2.0, and two vendor-specific licenses, recorded per model in Appendix~\ref{app:status}.

We decode every model deterministically, with temperature $0$, top-$p$ $1.0$, and a maximum of 128 new tokens. The integration check instead permits 16 new tokens, and DeepSeek-R1-Distill-Qwen-14B permits 512. The two dose-calibrated models run in bfloat16 and the integration check in float16. The full eight-model roster was screened for bfloat16 execution on a single 48-gigabyte accelerator. A fixed system prompt instructs every model to act as a careful introspection assistant and return only the requested JSON object. Two models need separate interpretation, for the same underlying reason. Our configuration does not disable the thinking segment that opens Qwen3-14B responses, and DeepSeek-R1-Distill-Qwen-14B opens with a reasoning chain that can run past its 512-token budget. In both cases the introspective probe may be scoring reasoning tokens rather than a final answer, which limits what a comparison with the rest of the roster can mean.

The experiment pins an immutable revision for every model it runs, and Appendix~\ref{app:status} lists the revisions. A LoRA fine-tune of Qwen2.5-7B-Instruct serves as the known positive. It trained on 400 directions and is scored on 100 held-out directions disjoint from the training split, and Section~\ref{sec:res-sensitivity} reports the held-out result. The adapter is a LoRA fine-tune of Qwen2.5-7B-Instruct at revision \texttt{a09a3545}, trained for three epochs on 800 rows, 400 with an intervention present and 400 sham, with rank 8, scaling 16, and adapters on all seven attention and feed-forward projections. We trained the adapter ourselves, so it carries no upstream revision identifier, and the adapter weights are the immutable artifact we release in its place.

\subsection{Benchmarks}

We test reportability across twelve benchmarks that span computations relevant to oversight. MMLU \citep{hendrycks2021mmlu} and MMLU-Pro \citep{wang2024mmlupro} cover general knowledge; HellaSwag \citep{zellers2019hellaswag}, WinoGrande \citep{sakaguchi2020winogrande}, and ARC-Challenge \citep{clark2018arc} cover commonsense; and GPQA \citep{rein2023gpqa} covers graduate-level science. GSM8K \citep{cobbe2021gsm8k} and MATH \citep{hendrycks2021math} cover arithmetic and mathematical reasoning; TruthfulQA \citep{lin2022truthfulqa} covers truthfulness; HumanEval \citep{chen2021humaneval} and MBPP \citep{austin2021mbpp} cover program synthesis; and IFEval \citep{zhou2023ifeval} covers instruction following. They differ in what a correct answer requires. Whether a difference in reportability across them reflects the computation probed or the surface of the task is an empirical question rather than something the selection settles, and we treat it as one throughout: domain and answer format vary together in this suite, so the two cannot be separated by benchmark choice alone. Answer format varies as well. The multiple-choice benchmarks hold scoring constant while the underlying computation changes. The free-text and code benchmarks then vary the format, so answer format alone cannot explain a profile difference. Table~\ref{tab:benchmarks} gives the per-benchmark item counts actually sampled in each battery.

\begin{table}[t]
  \centering
  \scriptsize
  \setlength{\tabcolsep}{3.5pt}
  \caption{Implemented benchmark suite and distinct items sampled per benchmark in each battery. The dose battery samples 32 items per benchmark on two models, the eight-model battery samples 2 per benchmark on all eight models, and the breadth battery samples 4 on each of four benchmarks across three models. A dash indicates that the battery does not use the benchmark. We read item counts from scored artifacts rather than the run specification, so they record completed measurements.}
  \label{tab:benchmarks}
  \begin{tabular}{llcccl}
    \toprule
    Benchmark & Domain & Dose & Nine-model & Breadth & Source, configuration, split \\
    \midrule
    MMLU \citep{hendrycks2021mmlu} & General knowledge & 32 & 2 & 4 & cais/mmlu, all, test \\
    MMLU-Pro \citep{wang2024mmlupro} & General knowledge & 32 & 2 & -- & TIGER-Lab/MMLU-Pro, default, validation \\
    HellaSwag \citep{zellers2019hellaswag} & Commonsense & 32 & 2 & -- & Rowan/hellaswag, validation \\
    WinoGrande \citep{sakaguchi2020winogrande} & Commonsense coreference & 32 & 2 & 4 & allenai/winogrande, debiased, validation \\
    ARC-Challenge \citep{clark2018arc} & Science commonsense & 32 & 2 & 4 & allenai/ai2\_arc, ARC-Challenge, test \\
    GPQA \citep{rein2023gpqa} & Graduate-level science & 32 & 2 & -- & Wanfq/gpqa, gpqa\_main, train \\
    GSM8K \citep{cobbe2021gsm8k} & Grade-school arithmetic & 32 & 2 & 4 & openai/gsm8k, main, test \\
    MATH \citep{hendrycks2021math} & Mathematical reasoning & 32 & 2 & -- & HuggingFaceH4/MATH, default, test \\
    TruthfulQA \citep{lin2022truthfulqa} & Truthfulness & 32 & 2 & -- & EleutherAI/truthful\_qa\_mc, validation \\
    HumanEval \citep{chen2021humaneval} & Program synthesis & 32 & 2 & -- & openai/openai\_humaneval, test \\
    MBPP \citep{austin2021mbpp} & Program synthesis & 32 & 2 & -- & huggy-1/mbpp, test \\
    IFEval \citep{zhou2023ifeval} & Instruction following & 32 & 2 & -- & google/IFEval, default, train \\
    \bottomrule
  \end{tabular}
\end{table}

Benchmark licenses are recorded in Appendix~\ref{app:status}.

MMLU items are four-option multiple-choice questions spanning 57 subjects, and WinoGrande items are sentences with a blank and two candidate fillers; both expect one option letter as the answer. For these two benchmarks, the implemented scoring coerces the model response to an option letter and applies an exact 0/1 match to the gold letter. GSM8K items are natural-language word problems of grade-school arithmetic reasoning that expect a brief free-text final answer, and the implemented scoring takes the maximum of normalized exact match and token F1 against the final answer extracted after \texttt{\#\#\#\#}.

The instrument-validation study draws on fixed copies of benchmark items rather than retrieving them afresh, so it is reproducible from the artifacts alone. It draws two items from each of three files. Per-item provenance traces MMLU to the Hendrycks test repository, WinoGrande to the AllenAI WinoGrande dataset, and GSM8K to the original test data. The conversions and scoring rules stated above then apply unchanged. The validation study therefore uses six source items in total, while the evaluation samples at most 32 items from each benchmark.

Five questions organize the design. The first is whether a report tracks the intervention at all, measured as a detection margin over sham, and, if it does, whether it tracks what was perturbed rather than only that something was, measured as a reconstruction margin over the observer bound. The two together would separate sensitivity from access. The first is null here, so the conjunction cannot establish access. Reconstruction is separately unscored for a reason of its own, which Section~\ref{sec:res-domain} gives: these interventions carry no ground-truth concept label for a reconstruction to be right or wrong about. The dissociation we report is therefore drawn against the linear probe and the fine-tuned known positive rather than against reconstruction. The second is whether reportability has structure across the design: whether variance in the reportability profile concentrates at the level of task domain and computational object rather than at the level of the random seed. We quantify those components with intraclass correlations and interval estimates rather than cell-wise significance tests (Appendix~\ref{app:formal}). The third is whether reportability decays as the intervention recedes in time and as context is withheld. The three temporal tracks are designed to separate those two things, and this release does not execute a genuinely delayed path, so what we measure across tracks is context masking rather than delay (Section~\ref{sec:res-temporal} and Appendix~\ref{app:status}). The fourth is whether a report tracks damage to the output rather than the intervention itself, which is why the dose ladder places its evidential weight on doses that leave task accuracy unchanged. The fifth is how much of any apparent introspection is first-order leakage: the dose-response intercept gives the detection margin as output divergence approaches zero, and the gap between that intercept and the aggregate margin estimates the leakage share.

Calibration and control conditions tell us when the instrument is measuring guessing or artifact sensitivity. We treat those outcomes as manipulation checks. The primary analysis comprises the detection and reconstruction margins against the sham and observer conditions. For domain, layer, operator, and track analyses, we apply false-discovery control and hierarchical partial pooling across the design rather than independent cell-wise tests.

\section{Results}\label{sec:results}

\begin{figure}[t]
  \centering
  \includegraphics[width=\linewidth]{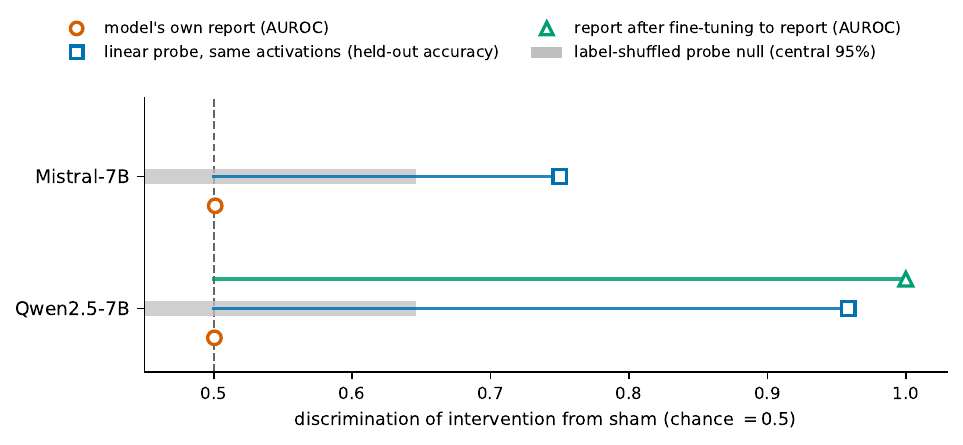}
  \caption{Three channels for the same intervention at the layer-16 residual-stream site, shown on scales with a chance level of 0.5. We score the model's report as AUROC over paired intervention and sham trials from the dose battery, the linear probe as held-out classification accuracy on activations captured at the same site, and the fine-tuned known positive as AUROC on 100 held-out directions absent from training. Although AUROC and accuracy are different statistics, each measures discrimination between intervention and sham with the same chance level. The grey band spans the central 95 percent of a 200-refit label-shuffled null for each linear probe; no permutation in either set reaches the measured margin. The reports remain at chance, while the linear probes recover the intervention and the fine-tuned model reports it.}
  \label{fig:dissociation}
\end{figure}

Three measurements tell the story. At the layer-16 residual-stream site under immediate probes, none of the eight models discriminates a controlled intervention from a matched sham. The same instrument gives near-ceiling discrimination to a model fine-tuned to report that class of intervention on held-out directions. A linear probe also recovers intervention presence well above chance from the same activations in both dose-calibrated models. The information is in the activations, but not in the reports. Figure~\ref{fig:dissociation} compares these three channels on a common chance scale.

\subsection{Instrument validation}\label{sec:validation}

We first validated the instrument on Qwen2.5-7B-Instruct and Mistral-7B-Instruct-v0.3. The validation study used six items from MMLU, WinoGrande, and GSM8K; one residual-stream site; one dose; the immediate track under both hook temporal scopes; two seeds; and paired intervention, sham, random-direction, and observer conditions. We scored all 192 measurements. These runs validate the procedure and estimate item-level variance. It supports no claim about any model, and we make none from it.

The validation runs show that the controls, report parser, and execution paths behave as intended. First, the controls produce the intended discrimination behavior. Qwen2.5-7B-Instruct reports no change on any probe regardless of condition. A raw yes-rate would register this as a constant non-detection floor. The paired sham design scores it exactly at chance ($d' = 0.0$, AUROC $= 0.500$ over 24 complete pairs), with confident reports whose calibration is uncorrelated with intervention presence (Brier $0.36$). Second, whether a model produces a scorable report at all is a measurement outcome, not a nuisance. Depending on condition, Mistral-7B-Instruct-v0.3 failed to produce a parseable introspective report on between 17\% and 33\% of runs. Both the complete-report and inclusive analyses of Appendix~\ref{app:analysis} score its remaining 20 complete pairs at chance ($d' = 0.0$, AUROC $= 0.500$). Third, neither model shows a detection margin over sham at this scale. We interpret none of it. With no dose ladder, no linear-probe anchor, and twenty to twenty-four pairs per model, these runs cannot separate absent reportability from an insensitive instrument. The full evaluation supplies exactly what that distinction needs.

After instrument validation, we ran an exemplar suite on pretrained checkpoints to exercise the remaining design axes. The suite covered Track B reflection with visible and masked context under both hook temporal scopes, and Track C spontaneous reflection. It also covered an attention-head intervention on the slice corresponding to one query head at a middle layer, a dose ladder with three levels at the residual-stream site, and an observer pairing in which a 0.5B-parameter target model is scored by a stronger 7B-parameter model. All six evaluation groups completed, producing complete records for 46 runs over 22 matched pairs. We then scored every artifact. The detection-family probes span 40 of the 46 artifacts, 20 true-intervention rows and 20 true-sham rows; recomputing each score reproduced the stored value exactly on all 40, and the self-reported detection rate was 0 of 20 on true-intervention rows and 0 of 20 on true-sham rows. Not once did the model report a detected alteration, whether or not one had occurred. The condition-blinded rubric scored the suite's two completed Track C pairs, drawn from the MMLU abstract-algebra and international-law items, and found a rubric-score delta of zero between the baseline and intervention reflections on both pairs. The suite's two stronger-model-observer rows both carry a true condition of intervention, so that arm of the observer bound has no matched sham row and cannot be computed here. A classifier observer trained on the suite's remaining rows scored at essentially chance: a margin of $-0.0013$, 53.8 percent held-out accuracy against a 53.8 percent chance baseline, over 13 held-out rows. With $n = 22$ pairs from two benchmark items and eleven intervention or track combinations, the suite validates execution, not model capability. Every axis runs end to end and every artifact reaches a working scorer. These data support no claim about detection, localization, or introspection.

\subsection{Sensitivity validation}\label{sec:res-sensitivity}

A detection null matters only if the signal exists and the instrument can register it. The linear probe tests whether the intervention leaves a linearly extractable trace in the activations. The known-positive model tests whether the full pipeline scores a model trained to report that trace.

For each site in the dose-calibrated evaluation, we train the linear probe of Appendix~\ref{app:formal} on held-out activations and report its accuracy margin $m_{\mathrm{probe}}$ over the chance level, which is a lower bound on the linearly available information rather than a ceiling; a site whose margin sits near zero drops out of the primary analysis, because a silent model tells us nothing at a site where the intervention left no linearly available trace. At the layer-16 residual-stream site the margin is far from zero for both dose-calibrated models. Each probe trains on 144 activation vectors and is scored on 48 held out, drawn from 96 items run through paired sham and intervention forward passes and mean-pooled over prompt positions. On Qwen2.5-7B-Instruct the probe reaches 0.958 held-out accuracy, a margin of $m_{\mathrm{probe}} = 0.458$ over the 0.500 chance level. On Mistral-7B-Instruct-v0.3 it reaches 0.750, a margin of $0.250$. We pair each probe with a label-shuffled permutation control. Because the probe margin forms one side of the central dissociation, we estimate its null with 200 refits per model. After label shuffling and retraining, the margin distribution is centered near zero: $-0.005$ with a standard deviation of $0.073$ for Qwen2.5-7B-Instruct, and $-0.007$ with a standard deviation of $0.075$ for Mistral-7B-Instruct-v0.3. No permutation in either set reaches the measured margin, which places both at $p \le 0.005$ under this control, and the measured margins stand $6.3$ and $3.3$ null standard deviations above zero. The 95th percentile of each null sits at $0.125$. A margin of that size is what label shuffling alone produces at this held-out size, and it is evidence of nothing.

The release's pass criterion for this control is mis-specified, and we report its failure rather than quietly drop it. The criterion compares the largest absolute permutation margin against a fixed threshold of $0.1$. A sampled maximum grows with the number of draws, so the better the null estimate becomes, the harder the test is to pass, which is backwards. It fails at 5 permutations, where the largest absolute margins are $0.125$ and $0.104$, and by a larger margin at 200, where they are $0.229$ and $0.208$. The quantity it should compare is the measured margin against a quantile of the null, which is the test reported above and which both probes pass comfortably. We leave the flag standing in the released artifacts and record here that it does not bear on the result.

These two margins were measured at layer 16, where the intervention acts, rather than downstream of it where a report would have to be formed. We therefore re-harvested the same probe at deeper capture layers, holding the intervention at layer 16 and the seed fixed so that only the read-out position changes. The trace does not decay downstream; it sharpens. Reading at layers 20 and 24, both models separate intervention from sham with no held-out error, and Mistral-7B-Instruct-v0.3 does the same at layer 31. Qwen2.5-7B-Instruct has 28 layers, so a layer-31 read-out does not exist for it. Each downstream probe carries its own 200-refit label-shuffled control, and no permutation reaches the measured margin in any of them. Perfect separation on a 48-item held-out split states an absence of errors rather than an accuracy estimate, and we report it as such. The intervention is not merely traceable at the site we perturbed; it remains fully linearly decodable at every downstream depth we tested, including the last layers before the model writes its answer, which is where a report about the change would have to originate.

The steering-aware fine-tuned model serves as the end-to-end known positive. That such a model can be trained is established rather than novel: \citet{fonseca2025steeringawareness} report high detection accuracy on held-out concepts with no false positives on clean controls across seven models, and \citet{shenoy2026introspectionadapters} train models to report implanted behaviors. We reproduce the effect inside our own pipeline so that the instrument, and not only the capability, is what the result validates. We fine-tuned Qwen2.5-7B-Instruct with a LoRA adapter on 400 training directions, seeds 500000 through 500399, and scored it on 100 held-out directions, seeds 900000 through 900099, that never appeared during training. The run specification enforces this disjointness by construction and verifies it again at evaluation time. The evaluated site is the residual stream at layer 16, with the intervention replacing the activation with a random reference direction at token positions 0 through 2. On this held-out set, the model reported "detected" on 100 of 100 intervention trials and "not detected" on 100 of 100 sham trials, with zero parse failures across the 200 scored trials ($d' = 5.1517$, AUROC $\approx 1.0$). Zero empirical errors at $n = 100$ pairs bounds the true error rate at roughly under 3 percent with 95 percent confidence; it does not establish an exactly zero error rate. The reported confidence interval collapses to a single point at this data pattern, because resampling data with no observed variance always returns the same value. That collapse should be read against the error-rate bound above, not as proof of a literally error-free instrument. This evaluation draws its intervention and sham prompts independently from the same item pool rather than matching them item by item, unlike the sham design used throughout the rest of this section. The known positive establishes that OWMI registers a strong signal at the trained intervention site when a model has learned to report it. This validates instrument sensitivity, but it does not test introspection in the eight roster models of Section~\ref{sec:res-detection}.

\subsection{Detection and discrimination}\label{sec:res-detection}

For the primary detection contrast, we report margins over sham by model and benchmark as $d'$ and AUROC with item-pair clustered bootstrap intervals. We also report margins over the observer and random-direction controls. No model clears the sham control. The reconstruction family carries no estimate here, and Section~\ref{sec:res-domain} gives the reason: the interventions we run supply no ground-truth label for the altered concept, so there is nothing for a reconstruction report to be scored against.

\begin{figure}[t]
  \centering
  \includegraphics[width=\linewidth]{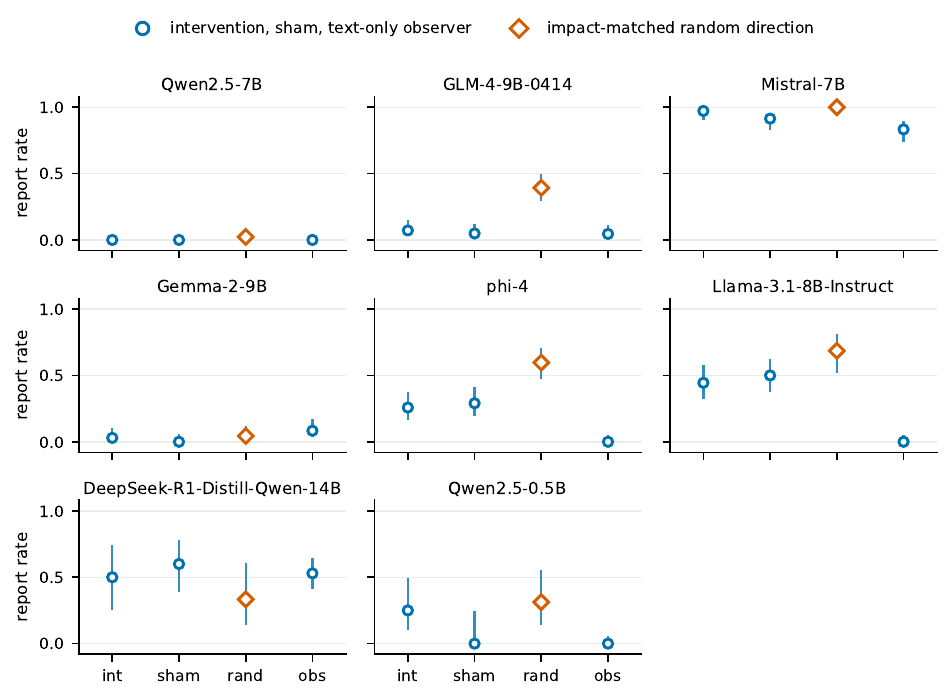}
  \caption{Raw self-report rate for eight models across four conditions at the layer-16 residual-stream site. The rate is the share of parseable trials whose probe answer decodes to a detected change. Each panel contains one model because the relevant comparison is across conditions within a model. Error bars are 95 percent Wilson intervals. Conditions are intervention, sham, impact-matched random direction, and text-only observer; diamonds mark the random-direction control. Report rates vary widely across models but little between intervention and paired sham within each model. The random-direction rate exceeds the intervention rate in seven of the eight models with interpretable rates. Figure~\ref{fig:dprime} scores detection as the paired contrast against sham.}
  \label{fig:report-rate}
\end{figure}

Track A measurements cover all eight models, twelve benchmarks, the layer-16 residual-stream site, and both hook temporal scopes. This scope represents one site and one dose level. Seven models use unit-norm random-direction controls and two seeds; Qwen2.5-7B-Instruct and Mistral-7B-Instruct-v0.3 use the calibrated dose ladder of Section~\ref{sec:res-dose}. Every model has returned results across the benchmark-by-scope cells at this site and all 216 cells returned results. Complete cell coverage does not guarantee a scorable pair in every cell. What empties the remaining cells is parse failure rather than a missing run: DeepSeek-R1-Distill-Qwen-14B's MMLU cells and Llama-3.1-8B-Instruct's GSM8K cells return zero complete pairs because every trial in those cells fails to parse, and Figure~\ref{fig:report-rate} plots the raw self-report rate for each model with a scorable pair under each condition, and Table~\ref{tab:detection-margins} reports the paired detection contrast against sham for all eight models, together with the margin over the observer and random-direction controls.

\begin{figure}[t]
  \centering
  \includegraphics[width=\linewidth]{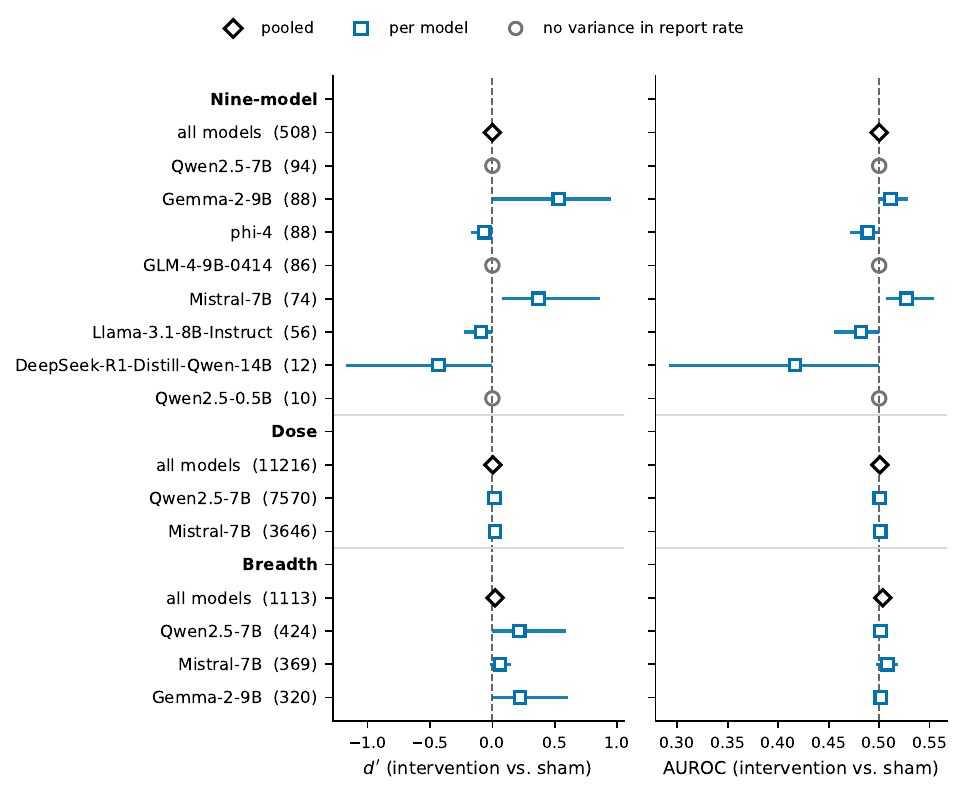}
  \caption{Detection contrast between intervention and paired sham conditions, per model and pooled, in both batteries at the layer-16 residual-stream site, with 95 percent item-pair clustered bootstrap intervals. The number in parentheses is the count of complete intervention-sham pairs. Both scales are shown because at these sample sizes they carry different news: an interval can exclude zero on $d'$ while the corresponding AUROC sits within a thousandth of chance. In the dose battery the intervals are narrower than the marker, and the discrimination they bound is negligible in size (Section~\ref{sec:res-detection}). In the eight-model battery the per-model estimates scatter widely, and that scatter tracks pair count rather than any property of the models: the two widest positive intervals belong to Gemma-2-9B-IT and Mistral-7B-Instruct-v0.3 on 88 and 74 pairs, against 11{,}216 pairs for the pooled dose estimate. Both batteries are complete, and the scatter did not resolve as measurements accumulated: Mistral-7B-Instruct-v0.3 moved from $d' = 0.538$ on 68 pairs to $0.370$ on 74 as the battery closed, which is the behaviour of an unstable estimate rather than a converging one. Open grey circles mark cells whose report rate does not vary at all, so the interval has zero width; the cross marks a cell with no complete pair.}
  \label{fig:dprime}
\end{figure}

No model clears the sham control, and none clears the random-direction control either. The strength of that second statement differs across the roster. Qwen2.5-7B-Instruct and Mistral-7B-Instruct-v0.3 carry a calibrated control, and the two calibrations did not succeed equally well. The search looks for a random-direction strength whose mean Jensen-Shannon divergence from the baseline next-token distribution matches the treatment's. For Qwen2.5-7B-Instruct it landed within about one percent of target, at a divergence of $0.231$ against the treatment's $0.234$, which is impact matching in the sense the term should carry. For Mistral-7B-Instruct-v0.3 it landed about forty percent above target, at $0.080$ against $0.057$, so its random direction does more damage to the output than the intervention does. The two perturbations are therefore not of equal impact for this model, and the direction of the mismatch does not make the comparison safe in some compensating way: it means the intervention-versus-random contrast asks whether the model separates its targeted intervention from a perturbation that disturbs the output more, which is a different question from the one impact matching is meant to pose. Only Qwen2.5-7B-Instruct carries an impact-matched random-direction control; Mistral-7B-Instruct-v0.3 does not. The remaining six models in this battery carry a unit-norm random direction, which equates the size of the perturbation but not its effect, and which is therefore the weakest of the three and the one prior work also uses. Table~\ref{tab:detection-margins} reports the per-model contrasts and Figure~\ref{fig:dprime} their intervals. For Qwen2.5-7B-Instruct, the model does not separate the object we targeted from a random perturbation of equal downstream impact. For Mistral-7B-Instruct-v0.3 the comparison is against a perturbation of roughly 40 percent greater impact, and for the other six against a perturbation of equal norm.

Here the magnitude matters more than the sign. With a campaign this large, an interval excluding zero can surround an effect too small to support detection. Pooling the 11{,}216 complete pairs of the dose battery, the detection contrast is $d' = 0.0039$ with a 95 percent interval of $[0.0017, 0.0062]$ and an AUROC of $\approx 0.5007$. The per-model estimates behave the same way: $d' = 0.017$ for Qwen2.5-7B-Instruct over 7{,}570 pairs and $d' = 0.022$ for Mistral-7B-Instruct-v0.3 over 3{,}646, at AUROCs of $0.5005$ and $0.5011$.

The equivalence test of Section~\ref{sec:equivalence} makes the bound explicit. Inverting the two one-sided tests, the data place the discrimination advantage below $0.15$ percentage points of AUROC at $p < 0.0001$, below $0.125$ percentage points at $p < 0.01$, and below $0.11$ percentage points at $p < 0.05$. A tenth of a percentage point is the first margin the data cannot exclude, at $p = 0.081$, which places the resolution limit of this estimate just under that value. The estimate resolves discrimination advantages down to about one tenth of an AUROC percentage point. Would a model whose reports discriminate at an AUROC of $50.1$ percent be introspecting? It would order one additional case correctly in every thousand, and no monitoring scheme could act on a signal that thin. The interesting boundary is not where our resolution runs out but where reportability would become usable, and that boundary sits far above anything these models produce. The effect is not merely unproven; it is unequivocally bounded into a range where no discrimination advantage could matter operationally. We read none of these estimates as above chance. An AUROC of $\approx 0.5007$ is an advantage of seven ten-thousandths, and a single defensible scoring choice moves the same quantity by more than that. One defensible alternative codes an unparseable answer as zero report evidence instead of dropping its pair. That change moves the pooled estimate from $+0.0039$ to $-0.1217$. The resulting shift of $0.127$ is thirty-two times the point estimate and reverses its sign. An estimate that one scoring decision moves by thirty-two times its own magnitude cannot be read as report discrimination. The estimand as scored mixes discrimination with condition-dependent parseability, and nothing in these data separates the two. At $n$ in the thousands, an interval narrow enough to exclude zero reports the precision of the estimate rather than the existence of an effect.

The detection result is therefore precisely estimated, negligible in size, and sensitive to a scoring choice at every sample size we tested. It establishes neither exact absence nor positive detection. We report the intervals as measures of precision, not as evidence that the estimates differ meaningfully from zero.

The eight-model battery, which carries far fewer pairs per model, pools to $d' = 0.0000$ with an interval of $[-0.041, 0.043]$ over 508 pairs. Two of its per-model cells carry positive point estimates, Gemma-2-9B-IT at $d' = 0.531$ over 88 pairs and Mistral-7B-Instruct-v0.3 at $d' = 0.370$ over 74, alongside negative estimates of similar magnitude for DeepSeek-R1-Distill-Qwen-14B and Llama-3.1-8B-Instruct. We do not base a claim on these cells, and the way they moved as the battery closed is the reason. Mistral-7B-Instruct-v0.3 stood at $d' = 0.538$ on 68 pairs before the last measurements landed and at $0.370$ on 74 after, a shift larger than most of the effect it was reporting. The same models sit at chance wherever the sample is large: Mistral-7B-Instruct-v0.3 scores $d' = 0.022$ over 3{,}646 pairs in the dose battery and $d' = +0.061$ over 369 pairs in the breadth battery. We report the smaller cells for completeness.

Report production limits the interpretation of two models in the roster. Qwen2.5-0.5B-Instruct fails to return a scorable report on nearly every trial. DeepSeek-R1-Distill-Qwen-14B loses two-thirds of its intervention trials because its reasoning chain exceeds the response budget. A further model, Qwen3-14B, was run under the same protocol and could not be elicited from at all; we exclude it from the roster and report it in Appendix~\ref{app:excluded} rather than alongside models measured on the same terms. Its surviving pairs give a point estimate below chance, but the interval reaches chance and the surviving sample is selected, so no directional claim follows.

The likely cause in both 14B models is a thinking segment that the current configuration does not disable (Section~\ref{sec:plan}).

Two models differ sharply in raw reporting behavior even though their discrimination scores agree with the rest of the roster. Qwen2.5-7B-Instruct's parseable reports say no change occurred on essentially every trial in every condition, at rates of 0.000 under intervention, 0.000 under sham, 0.042 under the random direction, and 0.000 under the observer. Mistral-7B-Instruct-v0.3's parseable reports say a change occurred on nearly every trial, at 0.949, 0.895, 1.000, and 0.826 across those same four conditions, with 17 to 33 percent of its trials failing to parse depending on condition. Both patterns land at chance for one reason: neither model's report rate moves between intervention and sham. A response that never changes across conditions carries no information about the condition, and it makes no difference whether the constant answer is yes or no.

Three models have a nonzero paired margin against the random-direction control while remaining at chance against sham. All three margins are negative: the graded detection score is higher under the random perturbation than under the targeted intervention (Table~\ref{tab:detection-margins}). We read this as report production responding to representational disturbance rather than as access to the targeted object, since none of the three models clears either control. The raw rates point the same way and more broadly. In seven of the eight models whose random-direction and intervention rates are both interpretable, the model reports a change more often under an impact-matched random perturbation than under the targeted intervention (Figure~\ref{fig:report-rate}), and in some the gap is large: 0.580 against 0.295 for phi-4, 0.380 against 0.067 for GLM-4-9B-0414, and 0.711 against 0.500 for Llama-3.1-8B-Instruct. DeepSeek-R1-Distill-Qwen-14B is the single exception, on a sample thinned by parse failure. Because these rates are unpaired, they provide weaker evidence than the paired margins. They point in the same direction: report production responds to disturbance, not to the identity of the disturbed object. One further nonzero entry rests on a single complete pair and is not interpretable.

All eight models reach the same outcome at this site: no report discriminates the intervention from a matched sham. The three nonzero margins against another control point opposite to detection. Yet activations from the two directly probed models carry a linearly decodable intervention signal (Section~\ref{sec:res-sensitivity}). Under immediate probes at one residual-stream site and one dose level for seven of the eight models, the activations contain the signal and the reports do not.

\begin{table}[t]
  \centering
  \footnotesize
  \setlength{\tabcolsep}{4.2pt}
  \caption{Detection margins for all eight models, one residual-stream site, both hook temporal scopes pooled. Pairs is the number of complete intervention-sham pairs entering the paired discrimination estimate. $d'$ and AUROC score discrimination between the intervention and paired sham conditions (Equation~\ref{eq:dprime}); three of the eight models with at least one complete pair sit exactly at the chance values of 0 and 0.500, and the remainder scatter on both sides of chance on samples that parse failure has thinned, which Figure~\ref{fig:dprime} shows with the corresponding item-pair clustered bootstrap intervals; the excluded model Qwen3-14B is not shown (Appendix~\ref{app:excluded}). Of its 384 trials, 5 produced a scorable response and none formed a complete intervention-sham pair, so every cell would be either undefined or, in the case of its random-direction margin, a single observation. We report it in the text instead of giving it a row. Parse failures is the share of probe responses that did not decode into a scorable report, given as a range across the intervention, sham, and random-direction conditions. $\mu^{\mathrm{obs}}$ and $\mu^{\mathrm{rand}}$ are the mean paired differences in the graded detection score against the text-only observer and impact-matched random-direction controls respectively (Appendix~\ref{app:formal}), each parenthesized with the count of complete pairs entering that estimate; a margin of 0 means the report did not distinguish the two conditions, and insufficient pairs means no pair had both conditions scorable. All eight models now have complete six-cell coverage (Section~\ref{sec:res-detection}).}
  \label{tab:detection-margins}
  \begin{tabular}{lrrrrrr}
    \toprule
    Model & Pairs & $d'$ & AUROC & Parse failures & $\mu^{\mathrm{obs}}$ & $\mu^{\mathrm{rand}}$ \\
    \midrule
    Qwen2.5-0.5B-Instruct & 10 & $+0.000$ & 0.500 & 83--88\% & 0.333 ($n$=6) & -0.071 ($n$=14) \\
    Mistral-7B-Instruct-v0.3 & 74 & $+0.370$ & 0.527 & 19--30\% & 0.081 ($n$=74) & -0.030 ($n$=66) \\
    Qwen2.5-7B-Instruct & 94 & $+0.000$ & 0.500 & 0--2\% & 0.000 ($n$=94) & -0.043 ($n$=93) \\
    Llama-3.1-8B-Instruct & 56 & $-0.090$ & 0.482 & 29--53\% & 0.500 ($n$=64) & -0.235 ($n$=34) \\
    Gemma-2-9B-IT & 88 & $+0.531$ & 0.511 & 5--8\% & -0.067 ($n$=90) & -0.045 ($n$=89) \\
    GLM-4-9B-0414 & 86 & $+0.000$ & 0.500 & 4--8\% & 0.022 ($n$=90) & -0.295 ($n$=88) \\
    Phi-4 & 88 & $-0.065$ & 0.489 & 8--8\% & 0.295 ($n$=88) & -0.284 ($n$=88) \\
    DeepSeek-R1-Distill-Qwen-14B & 12 & $-0.431$ & 0.417 & 79--88\% & 0.167 ($n$=12) & 0.200 ($n$=10) \\

    \bottomrule
  \end{tabular}
  \vspace{2pt}

  \raggedright\footnotesize $^{\dagger}$Near-total parse failure leaves too few complete pairs for interpretation; reported for completeness only. $^{\ddagger}$Every sham trial fails to parse (96 of 96), and only 4 of 96 intervention trials parse; with zero complete intervention-sham pairs, $d'$ and AUROC are not defined. The random-direction margin shown rests on a single complete pair, the only pair anywhere in this model's conditions that parsed on both sides, and is not interpretable (Section~\ref{sec:res-detection}). $^{\S}$All six cells have returned results, but a majority of trials fail to parse because this model's reasoning chain can exceed the response budget (Section~\ref{sec:plan}); the pairs reported here are the subset that produced a scorable report in both paired conditions.
\end{table}

\subsection{Dose-response and the leakage bound}\label{sec:res-dose}

The dose ladder exists to separate a report that tracks the intervention from a report that tracks the damage the intervention does to the output. Crossing the ladder with measured output divergence gives the hierarchical dose-response model of Appendix~\ref{app:formal}, and from it the zero-divergence margin $m_0$ of Equation~\ref{eq:m0} and the leakage share $\lambda = 1 - m_0 / \bar{m}$.

We report neither quantity, and the two reasons are different in kind. The first is a limit of this release. The leakage share is a ratio whose denominator is the aggregate detection margin $\bar{m}$, a difference in report rates between the intervention and sham arms. Over the 11{,}216 complete pairs of the dose battery that difference is $0.0014$, the report-rate margin corresponding to an AUROC of $0.5007$. A ratio whose denominator is compatible with zero is not identified: the estimate is governed by noise in the denominator rather than by the mechanism the ratio isolates, and its interval is uninformative rather than merely wide. This is a property of the measurement rather than of the estimator, and it would hold for any leakage share computed against a null aggregate margin.

The second reason is an execution shortfall. Equation~\ref{eq:m0} defines $m_0$ over the used-object stratum, the runs in which the intervened object actually participated in the answer, and that stratification comes from the per-run causal-relevance patching check. That component is present in the release without a targeted test (Table~\ref{tab:status}) and has not scored the dose battery. The stratum the estimand is defined over therefore does not yet exist in our data. Fitting the dose-response model to the unstratified pairs would produce a number, but not the estimand, and we decline to report the one in place of the other.

Neither reason is an argument that $m_0$ is uninformative in general. It is the quantity that carries the evidential weight of any access claim, and against a system whose reports do clear the sham control it is the analysis that matters. The estimator is specified here in full and the dose ladder it consumes is released with the rest of the framework; the causal-relevance stratification it needs is present in the release without a targeted test, and validating it is the first step toward running this analysis anywhere. We leave the leakage decomposition for this roster open.

\subsection{Domain and object structure}\label{sec:res-domain}

The detection nulls of Section~\ref{sec:res-detection} rest on one computational object at one depth under immediate probes, and a null at a single design point is weak evidence. A report channel might carry the intervention only for objects the model has words for, only at depths where the perturbation survives to the output, or only when the probe arrives while the intervention is still running. We had to check. We therefore ran a breadth battery of 19{,}520 measurements over Qwen2.5-7B-Instruct, Gemma-2-9B-IT and Mistral-7B-Instruct-v0.3 that crosses four of the five probe families (detection, localization, characterization, and confidence, the last scored as calibration) together with the Track C spontaneous-reflection probe, both executed temporal tracks, three classes of computational object, six layer depths, and four benchmark domains. Attention heads and trained Gemma Scope sparse-autoencoder features enter our study here for the first time, with no earlier measurements of our own on either. The reconstruction family is absent from every figure and estimate that follows, and the reason is a property of the interventions rather than an omission. Reconstruction asks the model to name the altered concept, and scoring it requires a ground-truth label for that concept drawn from a closed set. Zero-masking a residual-stream site, attenuating an attention head, or suppressing a sparse-autoencoder feature we have not independently interpreted supplies no such label: there is no concept for the model to be right or wrong about. Scoring this family would require interventions on features whose semantics are established in advance, together with the closed category set and the human validation that Appendix~\ref{app:analysis} specifies. We hold the family in the released software rather than reporting an estimate we cannot ground.

\begin{figure}[t]
  \centering
  \includegraphics[width=\linewidth]{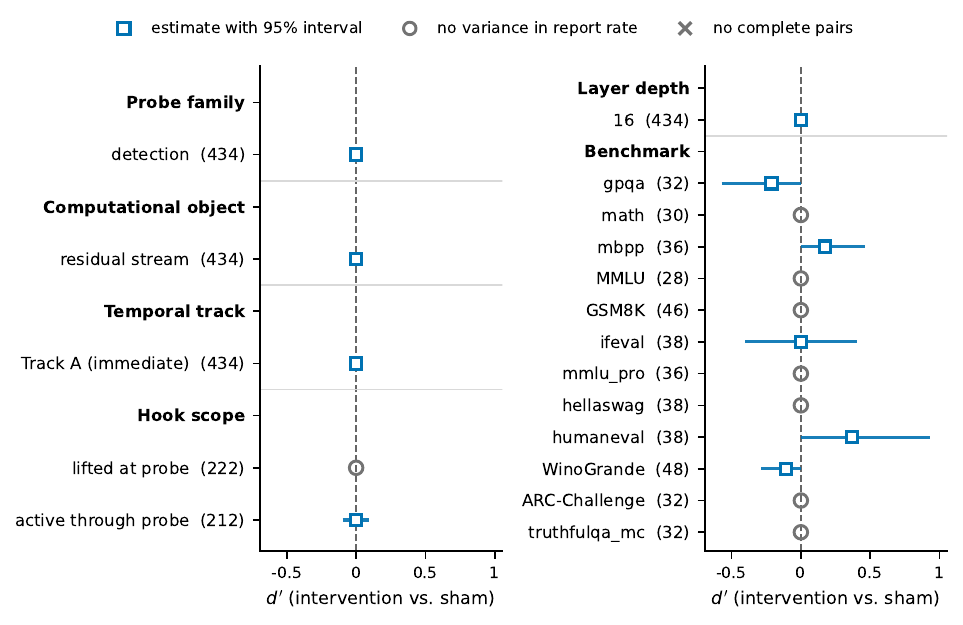}
  \caption{Paired intervention-versus-sham discrimination at every level of every design axis of the breadth battery, over 19{,}520 measurements on three models. Each block re-partitions the same 19{,}520 measurements along a different axis, so rows within the figure are overlapping views of one dataset rather than independent strata and their counts do not sum. Squares give $d'$ with 95 percent item-pair clustered bootstrap intervals; the number in parentheses after each level name is the count of complete intervention-sham pairs entering that estimate. Open grey circles mark cells in which the report rate does not vary at all, so the interval has zero width and records an absence of variation rather than a precisely estimated zero; the cross marks a cell with no complete pair. Every level's interval covers zero. The sparse-autoencoder object and the layer-20 depth carry the largest point estimates in the battery and are the two rows whose bars visibly extend to the right; both rest on the same sparse feature site, both have intervals reaching zero, and the AUROC of the sparse-autoencoder object is $0.5026$, so we read them as chance rather than as detection for the reason given in Section~\ref{sec:res-domain}. Attention-head and Gemma Scope sparse-autoencoder objects had no data at all before this battery. The localization row is a refusal rather than a null: models decline to name a layer in 967 of 1{,}184 intervention trials (Section~\ref{sec:res-domain}). Layer depths are residual-stream, attention-head and feature sites pooled within each depth.}
  \label{fig:axiscoverage}
\end{figure}

No level of any axis departs from chance in a way that changes the reading of Section~\ref{sec:res-detection} (Figure~\ref{fig:axiscoverage}). The detection probe family yields $d' = 0.022$ over 1{,}113 complete pairs, with an interval of $[0.000, 0.046]$ that reaches zero and an AUROC of $0.5036$. Every level's interval covers zero. Reports still fail to discriminate when we swap the residual-stream site for an attention head or a Gemma Scope feature, move the intervention from layer 8 to layer 31, or lift it before the probe arrives.

One cell needs both discrimination scales read together, and it is worth seeing why. The sparse-autoencoder object has the battery's largest point estimate, $d' = 0.232$ over 192 pairs, driven almost entirely by one feature site at layer 20 ($d' = 0.264$ over 64 pairs). Its interval reaches zero, and its AUROC is $0.5026$, within four ten-thousandths of chance. The measures differ because $d'$ contrasts two report rates on a probit scale, and the half-observation edge correction inflates that contrast when a small cell has a near-constant rate. AUROC ranks observations and remains near chance. Wherever a cell is sparse and its report rate barely moves, we read the AUROC and the interval rather than the $d'$ point estimate. On that reading the sparse-autoencoder object sits at chance with everything else.

Localization and Track C each need a separate reading (Figure~\ref{fig:axiscoverage}). Localization has no response variance to score. In 967 of 1{,}184 intervention trials the model says it cannot name a layer, 129 responses carry no scorable field, and only 88 offer a numeric guess, so models decline to localize in roughly four trials out of five. That is a fact about what the probe elicits, not a measurement of chance-level localization. Track C yields no complete intervention-sham pair from 1{,}344 measurements, so it validates execution coverage but provides no discrimination estimate. Wherever the report rate does not vary within a cell, the interval collapses to zero width; those cells carry no information about effect size and are drawn as such, because a zero-width interval records an absence of variation rather than a precisely estimated zero.

\subsection{Temporal decay}\label{sec:res-temporal}

The decay question can be put only to the tracks that produced scorable data. Track A and the masked-context Track B variant both do; Track C does not, so only the first leg of the ordering is available. As the closing paragraph of this section states, that variant withholds context without imposing a delay, so the contrast below separates context masking from context availability and not delay from either.

The two scorable conditions are indistinguishable. Reports under the immediate probe discriminate at $d' = 0.017$ over 508 complete pairs, and reports under the masked-context variant at $d' = 0.026$ over 605, with intervals that overlap each other and cover zero in both cases. Decay is not tested by this comparison, and we do not report it as untested-and-absent. Two things prevent the test. The manipulation that would produce decay, a genuine delay between intervention and probe, is not executed in this release, so the contrast varies context rather than time. And both quantities sit at chance, so there is no margin at either end for a decline to be measured between. We therefore decline to read the ordering of two negligible estimates as evidence about temporal decay in either direction.

Track C contributes nothing to this comparison. Its 1{,}344 measurements produced no complete intervention-sham pair, because the spontaneous probe invites an unprompted remark rather than an answer to a question, so intervention and sham rows do not pair. That is a property of the probe design rather than of the models, and it is the reason the third leg of the ordering remains open.

The delayed track as implemented masks the context, withholding the task prompt and the model's own output from the probe, but it runs on the immediate execution path. A genuine delay, in which the intervention is applied, the task completes, and the probe arrives later in a separate turn, is not implemented in this release. The Track A against Track B contrast we report therefore separates context masking from context availability, and does not yet separate delay from either.

\subsection{Calibration and report production}\label{sec:res-calibration}

We score confidence with the Brier score, equal-mass ECE, and selective prediction, and we report parse rates by cell and condition as defined in Appendix~\ref{app:analysis}. Figure~\ref{fig:calibration} compares the three confidence measures across all scored intervention and sham reports from the eight models.

Here, and only here, a self-report channel carries information about the intervention. In two of the eight models the confidence attached to a report predicts whether that report is correct better than chance: Qwen2.5-7B-Instruct reaches a selective-prediction AUROC of $0.647$ $[0.577, 0.719]$ over 190 scored reports, and GLM-4-9B-0414 reaches $0.587$ $[0.522, 0.651]$ over 178. These two were selected as the largest of eight estimates without a multiplicity correction, so we treat the stronger case as established and the weaker as suggestive. No remaining model exceeds chance: phi-4 reaches $0.521$ with a 95 percent interval of $[0.473, 0.572]$.

For Qwen2.5-7B-Instruct that statistic is not about calibration at all. Its detection report is constant: it answers that no change occurred on all 94 intervention trials and all 96 sham trials that produced a scorable report. Introspective correctness is therefore not an independent quantity for this model. It equals the sham indicator exactly, so the selective-prediction AUROC is, identically, the discrimination of intervention from sham by verbalized confidence. Computed as that contrast directly it is the same number, $0.647$ $[0.572, 0.723]$, and the direction is that the model is less confident when it has been intervened upon, with mean confidence $0.620$ under intervention against $0.858$ under sham.

The consequence is a dissociation between two channels of the same self-report rather than between meta-level and first-order access. The discrete verbal answer to ``did anything change'' carries no information about the intervention, at $d' = 0$ and an AUROC of exactly $0.500$. The confidence attached to that same answer discriminates the two at $0.647$. The null we report is about the discrete report, which is the channel an oversight scheme reads when it asks a model what happened.

GLM-4-9B-0414 does not license the same reading. Its detection report is not constant, at 6 of 90 intervention trials and 4 of 88 sham trials reporting a change, so correctness and condition are close but not identical for it. Scored as condition discrimination its confidence reaches $0.571$ $[0.503, 0.638]$, a lower bound within four thousandths of chance. We therefore rest the channel dissociation on Qwen2.5-7B-Instruct alone and treat GLM-4-9B-0414 as consistent with it rather than as independent support.

We put the selective-prediction result through the same scoring-robustness check the detection estimate failed. Rescoring every model's confidence under the coding that retains unparseable reports as zero evidence leaves both models in place, at $0.653$ $[0.581, 0.721]$ for Qwen2.5-7B-Instruct and $0.578$ $[0.514, 0.638]$ for GLM-4-9B-0414, and moves no other model above chance. The same scoring decision reverses the detection estimate but leaves this one intact. One qualification attaches to both: these intervals resample individual scored reports, whereas the detection intervals cluster on item pairs, so the two are not constructed alike and should not be read against each other as if they were.

Under the same protocol, report production ranges from no parse failures for Qwen2.5-7B-Instruct to near-total failure for Qwen2.5-0.5B-Instruct, which returns a scorable report on roughly one trial in seven. We treat parseability as a measurement outcome in its own right. A model that cannot produce a scorable report under a fixed protocol limits every oversight scheme that depends on eliciting one.

\begin{figure}[t]
  \centering
  \includegraphics[width=\linewidth]{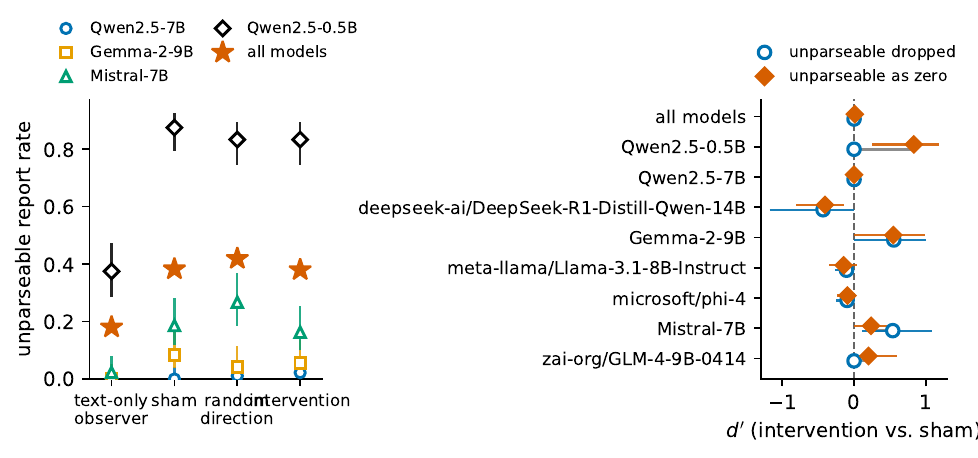}
  \caption{Differential parse failure and its effect on detection estimates in the 19{,}520-measurement breadth battery. Left: unparseable probe responses by condition, with 95 percent Wilson intervals. Among the three conditions that use the same reflexive probe, sham, random direction, and intervention, parse failure rises with disturbance. The intervention arm therefore loses more trials than its paired sham, selecting the surviving pairs. We include the text-only observer for completeness but exclude it from this comparison because its third-person probe differs in wording as well as disturbance. Gemma-2-9B-IT fails least often under the random-direction control, against the pattern of the other two models. Right: the paired detection contrast under both codings of an unparseable report. Circles drop the trial and its pair, the coding used elsewhere in this paper; diamonds retain the attempted report and credit it with no report evidence. Horizontal bars are 95 percent item-pair clustered bootstrap intervals. How far retaining unparseable trials moves an estimate tracks how asymmetric that model's parse failure is. Mistral-7B-Instruct-v0.3, whose intervention arm fails 6.7 percentage points more often than its sham, moves by $-0.250$; Qwen2.5-7B-Instruct, at a 2.7-point gap, moves by $-0.0004$; and Gemma-2-9B-IT, whose intervention arm actually parses slightly better than its sham, does not move at all. The correction bites exactly where the selection it corrects for exists.}
  \label{fig:parsesensitivity}
\end{figure}

Parse failure also biases the detection estimate in a specific direction. We compute each reported discrimination score only from pairs in which both conditions produce a scorable report, so parse failures remove trials from the analysis. Loss that ignored condition would cost us only sample size. The loss we observe does not ignore condition. Across the breadth battery the unparseable rate rises monotonically with how much the condition disturbs the model, among the conditions that share a probe: 11.1 percent under sham, 14.1 percent under an impact-matched random direction, and 14.4 percent under the intervention. The text-only observer sits well below all three at 7.3 percent, and we do not read that gap as part of the same trend, because the observer probe is reworded from the reflexive to the third person (Section~\ref{sec:controls}) and so differs from the other three in wording as well as in disturbance. The comparison that matters for the detection estimate is intervention against its paired sham, and those two share a probe exactly (Figure~\ref{fig:parsesensitivity}, left). Trials therefore go missing from the intervention arm more often than from its paired sham, which leaves the surviving pairs a selected subset rather than a random one.

We test this selection effect by rescoring every cell under a rule that retains each attempted report and assigns zero report evidence to failures. Under this rule, a model cannot benefit from omitting a response. The two codings bracket the estimate (Figure~\ref{fig:parsesensitivity}, right). Dropping unparseable trials yields a pooled detection contrast of $d' = 0.022$ with a 95 percent interval of $[0.000, 0.046]$, which reaches zero. Retaining them as zero evidence yields $d' = -0.061$ with an interval of $[-0.103, -0.022]$. Neither the positive nor the negative estimate is directional evidence. The negative shift comes from differential dropout, because the arm that fails more often receives more zeros. Between them the two codings show that the estimate does not survive this scoring decision, and neither sign supports a claim. The exclusion of unparseable reports was therefore mildly favorable to the models, and correcting for it does not recover a positive result: under neither coding does any cell rise above chance in the direction of detection. The per-model shifts are a check on the correction itself rather than a separate finding. They scale with each model's parse asymmetry: Mistral-7B-Instruct-v0.3 loses 6.7 more percentage points of its intervention arm than of its sham and its estimate moves by $-0.250$; Qwen2.5-7B-Instruct, at a 2.7-point gap, moves by $-0.0004$; and Gemma-2-9B-IT, whose intervention arm parses slightly better than its sham, does not move at all. A correction for selective dropout that moved estimates where no selective dropout exists would be doing something other than it claims. This one moves exactly where the selection is. The dose battery shows the same effect an order of magnitude larger, because it is an order of magnitude larger: over its 11{,}216 complete pairs the contrast moves from $d' = 0.0039$ $[0.0017, 0.0062]$ under the complete-pair coding to $d' = -0.122$ $[-0.134, -0.109]$ once unparseable trials are retained as zero evidence. We report the complete-pair coding throughout for comparability with the per-model tables, and note that the inclusive coding strengthens rather than weakens the conclusion.

We keep the inclusive coding secondary for one reason. When most trials fail to parse, it measures parse rate and calls the result discrimination. Qwen2.5-0.5B-Instruct illustrates the problem. It contributes 10 complete pairs, and the inclusive coding changes a constant report rate into $d' = 0.83$ $[0.25, 1.18]$, an apparent detection effect driven almost entirely by asymmetric parse failure. The inclusive estimate is informative when complete pairs are plentiful and parse rates are high, but it can generate artifacts otherwise.

\begin{figure}[t]
  \centering
  \includegraphics[width=\linewidth]{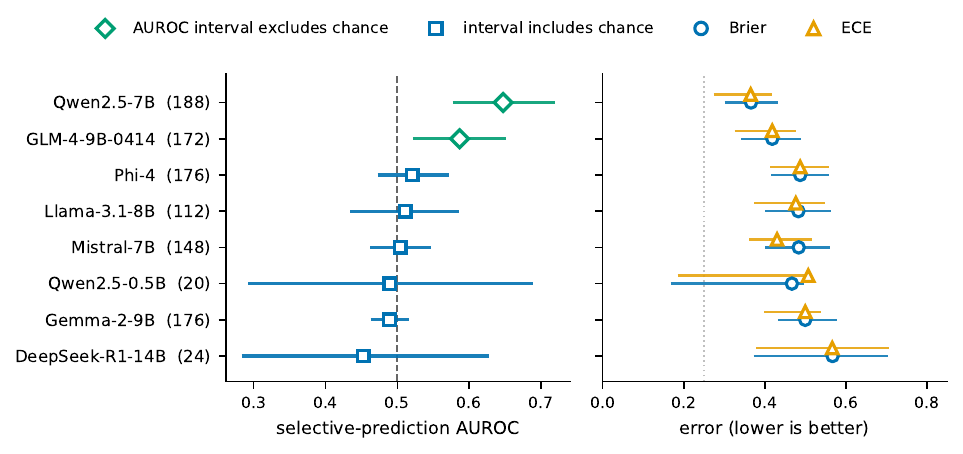}
  \caption{Confidence scoring for the eight-model battery, with discrimination and error measures in separate panels. Left: selective-prediction AUROC of verbalized confidence for introspective correctness, with 95 percent nonparametric bootstrap intervals over scored reports. The dashed line marks chance at 0.5, and green diamonds mark the two models whose intervals exclude chance. Right: Brier score and equal-mass expected calibration error on a shared scale, with lower values indicating better performance. We separate the panels because Brier score and expected calibration error are error measures, while AUROC measures discrimination. The count after each model name gives the number of scored reports and therefore reflects parse failure rather than a design target.}
  \label{fig:calibration}
\end{figure}

\section{Discussion}\label{sec:discussion}

Across eight open-weight models from seven laboratory families, no model distinguishes a controlled intervention on its computation from a matched sham by more than a negligible margin. The same instrument gives near-ceiling discrimination for a model fine-tuned to report that intervention. Chain-of-thought monitoring, self-critique, and other self-report oversight schemes assume that internal computation reaches a report. OWMI tests that assumption, and every tested model fails. Model behavior can differ between evaluation and deployment contexts \citep{ferrara2026defeat}, which sharpens the consequence: a monitoring scheme cannot assume that a model's account tracks the computation it describes.

Independent groups working in parallel report compatible results from different instruments: that the detection signal is confounded by a shift toward affirmative answers \citep{hahami2025disturbance}, that models cannot separate an intervention on their internal states from a manipulation of their input \citep{singh2026realitycheck}, and that the effect is indifferent to the content injected \citep{lederman2026contentagnostic}. What our measurements add is the form of the claim rather than its direction. An equivalence test converts a failure to detect an effect into a bound on how large the effect can be, and a bound is what an oversight argument needs. Reporting that a difference was not significant leaves open how much room the data still allow; reporting that discrimination sits below $0.15$ percentage points of AUROC does not.

The linear probe tells us where the failure sits: in the path from internal state to verbal report. At the intervention site, and with no held-out error at every downstream depth tested (Section~\ref{sec:res-sensitivity}), the intervention is linearly decodable from both dose-calibrated models at held-out accuracies of 95.8\% and 75.0\% against a 50\% chance level, while reports about the same event remain at chance. The information is in the internal state and does not reach the verbal report. An oversight scheme that reads activations works with a signal we have measured. One that asks the model has recovered none of it. The dissociation also hands training and prompting work a target with a scoreboard attached. We can measure whether an intervention builds the missing reporting path.

The negative result extends well beyond one site, but no farther than the measurements that support it. The eight-model battery covers twelve benchmarks at a single residual-stream site under immediate probes, at one dose level for seven of the eight models. The breadth battery adds three classes of computational object, including attention heads and Gemma Scope sparse-autoencoder features, across six layer depths, both executed temporal tracks, and four of the five probe families (detection, localization, characterization, and confidence, the last scored as calibration) together with the Track C spontaneous-reflection probe, on three of the models. What remains open is the delayed and spontaneous tracks at scale, the dose ladder beyond two models, models larger than fourteen billion parameters, and closed-weight systems. A null across these axes rules out nothing about introspective access in general. It does constrain any claim that such access exists for this class of internal event, at the sites and scales we measured.

Domain structure is the next measurement the instrument makes possible. A model that cannot report its arithmetic computation but can report truthfulness-relevant computation needs a different monitoring design from a model with the profile reversed.

OWMI measures one functional property, whether information about a controlled internal perturbation reaches the model's output channel. These results are about information flow. They say nothing about consciousness, experience, or moral status, and we intend no such reading.

\section{Limitations and Threats to Validity}\label{sec:limitations}

We measure the reportability of interventions we impose from outside. Generalizing from that to ordinary, unperturbed computation is a step this design supports only indirectly. A verbal report is behavior, not proof of privileged internal access, and our dissociation analyses bound the non-introspective explanations without eliminating them. The eight-model results cover immediate probes at one residual-stream site across twelve benchmarks, with one dose level for seven of the eight models, and the breadth battery extends three of those models to attention heads and sparse-autoencoder features over six depths, both executed tracks, and four of the five probe families (detection, localization, characterization, and confidence, the last scored as calibration) together with the Track C spontaneous-reflection probe. These nulls do not rule out introspective access at other sites, objects, doses, or tracks. The delayed and spontaneous tracks remain especially thinly covered. The measured population spans 0.5B to 15B open-weight parameters. Scale, post-training, and deployment conditions may all move the measured capability, so these profiles constrain hypotheses about frontier systems and do not estimate their reportability. An intervention can be strong enough to cause generic degradation rather than selective masking. The dose ladder turns strength into a measured variable, and it does not remove our choice of ladder range. Layer and object semantics differ across architectures, identical layer indices do not imply identical computational roles, and the localization probe inherits this limitation. Two constraints apply to the probe families not yet scored here. Automatic reconstruction metrics reward lexical overlap, which is why the observer condition sets the metric floor and why a human-validated subset must accompany any reconstruction claim. And spontaneous introspection is vulnerable to base-rate and prompting effects that sham-derived false-alarm rates bound only in part. The released software implements part of the full design, and Appendix~\ref{app:status} records the status of every component, without exception.

\section{Conclusions}\label{sec:conclusion}

Across eight open-weight models from seven laboratory families, no model's report of an intervention on its computation exceeds a matched sham by more than a negligible margin, and an equivalence test bounds that margin below 0.15 percentage points of AUROC: the reports are indistinguishable from random guessing. Linear probes recover the same intervention from the same activations at held-out accuracies of 95.8\% and 75.0\% against a 50\% chance level. The activations carry the information. The reports do not. The instrument detects the signal when a reporting path exists: a model fine-tuned for this class of intervention reaches $d' = 5.15$ and AUROC $\approx 1.0$ on held-out directions with the same pipeline. The failure therefore lies in the path from internal state to verbal report, not in the availability of information. Verbalized confidence produces a second dissociation, and it runs between channels rather than between levels. In Qwen2.5-7B-Instruct the discrete report discriminates at exactly chance while the confidence attached to it discriminates intervention from sham at $0.647$, so the signal reaches a graded quantity the model emits without reaching the words it chooses. For this class of internal event, oversight based on model accounts reads a channel without the signal, while activation-based oversight reads a channel that contains it. Chain-of-thought monitoring, self-critique, and confidence elicitation therefore need validating against an internal reference, never against the model's own testimony. OWMI is that reference, and we release it so that any claim of introspective access, present or future, has something to be scored against.

\bibliographystyle{unsrtnat}
\small
\bibliography{references}

\section*{Use of AI systems}
AI systems, including Claude (Sonnet, Opus, and Fable 5) and ChatGPT (Luna, Sol, and Terra 5.6), have been used at all stages of this research project.

\normalsize

\appendix

\section{Excluded Model and the Spontaneous Track}\label{app:excluded}

Two parts of the study ran to completion but are not evaluated on the same terms as the eight-model roster, and we separate them here rather than mixing them into estimates they cannot support.

\textbf{Qwen3-14B.} The model was run under the identical Track A protocol and returned a scorable response on 5 of 384 trials, forming no complete intervention-sham pair. The failure is in our elicitation rather than in the model's reporting: it emits a reasoning chain that the 128-token probe budget truncates before any answer is reached. Because no estimate can be formed, including it in a roster whose other members are measured on the same terms would misrepresent what was measured, so we exclude it. A re-run at a larger probe budget is the appropriate test and is not reported here. The same truncation mechanism affects DeepSeek-R1-Distill-Qwen-14B, which retains enough trials to remain in the roster with its parse-failure rate stated.

\textbf{Track C.} The spontaneous track ran 1{,}344 measurements in the breadth battery, balanced across intervention and sham conditions, and each run records the model's free-text reflection. It does not enter the reported estimates for two reasons. The paired estimator requires intervention and sham rows that join, and the track stores its baseline and intervened reflections within a single record, so no complete pair is formed. More substantively, the probe as implemented returns the model's reasoning about the benchmark item rather than an open-ended reflection on its own processing, so a null result here would speak to the probe design rather than to the models. Scoring the track would require a condition-blinded rubric validated against human annotation, because the distinction that matters is between a genuine remark about disturbed processing and ordinary task reasoning that happens to use the same vocabulary. A keyword rubric cannot make that distinction: applied to these reflections it flags the sham condition at the same rate as the intervened one, which is the signature of an instrument measuring vocabulary rather than detection. We release the track in the software and leave both the probe redesign and the annotation to future work.

\section{Release Scope and Component Status}\label{app:status}

Table~\ref{tab:status} separates what the measurement design specifies from what the current software release has actually run. Every listed component belongs to the protocol and exists in the software, and the status column records the strongest level we have established for it. ``Scored'' means that we ran the operation on pretrained checkpoints and report its results in Section~\ref{sec:results}. ``Validated'' means that a targeted test established the stated behavior. ``Executed'' means that we ran the operation on pretrained checkpoints without producing scored results. ``Implemented'' means that the operation exists without a targeted behavioral test. A declared setting counts for nothing here, and neither does a successful software load. Only a targeted test that established the stated behavior counts as validation.

The measurements reported in Section~\ref{sec:results} come from two complementary regions of the design rather than a single point. The eight-model and dose batteries hold the object class, site and track fixed, at the residual stream of layer 16 on the immediate track, and vary model, benchmark and dose. The breadth battery holds the model roster to three and varies the rest: four of the five probe families, both executed tracks, all three object classes including attention heads and trained Gemma Scope features, six sites across layers 8 to 31, and both hook temporal scopes. Two axes still carry no scored result. The reconstruction probe family did not execute, and Track C produced no complete intervention-sham pair, so neither contributes an estimate anywhere in this paper.

\begin{table}[t]
\centering
\small
\caption{Status of OWMI components in the current release. ``Validated'' requires a targeted behavioral test; execution alone does not qualify.}
\label{tab:status}
\begin{adjustbox}{max width=\linewidth, max totalheight=0.88\textheight}
\begin{tabular}{@{\hspace{1.1em}}P{0.33\linewidth}P{0.57\linewidth}}
\toprule
Component & Note \\
\midrule
    \rowcolor{owmiband}
    \multicolumn{2}{l}{\textbf{Scored} \;\footnotesize(produces an estimate reported in Section~\ref{sec:results})} \\[1pt]
    Benchmark adapters and canonical answer schemas & Twelve benchmarks in the eight-model and dose batteries, four in the breadth battery, with per-benchmark item counts in Table~\ref{tab:benchmarks} \\
    \rowcolor{owmirow}
    SAE-feature intervention & Single-feature delta substitution; run against trained Gemma Scope weights at layers 9, 20 and 31 in the breadth battery (Section~\ref{sec:res-domain}) \\
    Baseline/intervention task and probe runs & 3{,}072 runs across eight models; the detection results of Section~\ref{sec:res-detection} \\
    \rowcolor{owmirow}
    Sham condition & Same-item pairing throughout; 11{,}216 complete pairs in the dose battery and 1{,}113 in the breadth battery detection family \\
    Text-only observer condition & Produced the $\mu^{\mathrm{obs}}$ column for all eight models \\
    \rowcolor{owmirow}
    Per-run output-distribution divergence & Teacher-forced JS logging validated; materializes full-vocabulary distributions over every prompt position, which bounds measurable prompt length \\
    Paired sham d$'$/AUROC and calibration scoring & Produced every estimate in Section~\ref{sec:results} \\
    \rowcolor{owmirow}
    Random-direction controls & Deterministic validation passed; the eight-model battery used unit-norm controls, with calibrated norms on the two dose-calibrated models. Impact-matched norms have since been calibrated on a finer strength grid for all eight, and no reported battery uses them \\
    Immediate track (A) & 3{,}072 runs across eight models; the detection results of Section~\ref{sec:res-detection} \\
    \rowcolor{owmirow}
    Attention-head intervention & Scored on 212 complete pairs at four heads across layers 8 to 24 in the breadth battery (Section~\ref{sec:res-domain}) \\
    \rowcolor{owmirow}
    Delayed track (B), masked-context variant & Scored on 605 complete pairs in the breadth battery. Runs on the immediate execution path, so it withholds context rather than imposing a delay (Section~\ref{sec:res-temporal}) \\
    Run generation and cluster execution & Produced the runs of Section~\ref{sec:results} \\
    \addlinespace[2pt]
    \rowcolor{owmiband}
    \multicolumn{2}{l}{\textbf{Executed} \;\footnotesize(runs end to end, but produces no complete intervention-sham pair to score)} \\[1pt]
    Spontaneous track (C) & 1{,}344 measurements in the breadth battery. The probe invites an unprompted remark rather than an answer, so intervention and sham rows do not pair and no discrimination estimate is defined \\
    \rowcolor{owmirow}
    \addlinespace[2pt]
    \rowcolor{owmiband}
    \multicolumn{2}{l}{\textbf{Validated} \;\footnotesize(passes targeted tests; not exercised in the reported batteries)} \\[1pt]
    Hugging Face causal-language-model backend & Qwen2.5-0.5B GPU integration check passed \\
    \rowcolor{owmirow}
    Layer-level forward hooks & Residual and block path exercised in integration check \\
    Zero, scale, noise, and replacement operators & Prefill-only and shape-safe replacement regressions passed \\
    \rowcolor{owmirow}
    Hook temporal scope flag & Active-through-probe and lifted-at-probe paths both validated \\
    Stronger-model observer condition & Routing validation passed; reported observer margins use the same-model zero-shot tier, and the maximum over the observer family is not yet computed \\
    \rowcolor{owmirow}
    Masked-context Track B variant & Regression covers both probe paths and the context exclusions \\
    Compact run specification and shared model backend & Regression preserves sham, random-direction, and observer expansions \\
    \rowcolor{owmirow}
    Known-positive steering-aware fine-tune & LoRA on 400 held-in directions, scored on 100 disjoint held-out directions; $d' = 5.15$, zero parse failures over 200 trials \\
    Linear-probe sensitivity anchor & Trained on held-out activations at the evaluated site; held-out accuracies 95.8\% and 75.0\% against 50\% chance, with label-shuffled permutation controls \\
    \addlinespace[2pt]
    \rowcolor{owmiband}
    \multicolumn{2}{l}{\textbf{Implemented} \;\footnotesize(present in the release without a targeted test)} \\[1pt]
    Per-run causal-relevance patching check & Baseline replacement prompt-only; cannot fall through to random noise \\
  \bottomrule
\end{tabular}
\end{adjustbox}
\end{table}

The benchmark-adapter regression fixtures cover MMLU-Pro, HellaSwag, GPQA, HumanEval, MBPP, and IFEval. The attention-head intervention targets the query-head slice at the input of the attention output projection during prefill and supports grouped-query attention. Deterministic routing tests verify slice isolation and prefill-only firing on synthetic layouts, and the exemplar suite executes the intervention on a pretrained checkpoint at a mid-depth layer. The SAE-feature intervention performs prefill-only, single-feature delta substitution from user-supplied linear SAE weights. A row without a weights path raises an explicit error rather than silently falling back, and the release bundles no trained sparse autoencoder. The scoring component passed regressions for the common binary report, the finite edge-corrected probit, tied-score AUROC, item-pair clustered bootstrap intervals, and equal-mass ECE. The masked-context regression checks both probe paths and verifies exclusion of the task prompt and gold answer. The track component provides Track B reflection contexts with per-side outputs and the masked variant, together with Track C neutral reflection; the Track C rubric is implemented and has scored the exemplar suite's two completed pairs at zero delta, but it remains an unvalidated proxy against human judgment and has not yet scored the full battery.

The current software appends the introspection query to the benchmark prompt to construct a probe prompt. In the masked-context Track B condition, it instead supplies the probe without the task prompt, task output, or gold answer. We decode and grade only the generated continuation. Intervention functions modify declared prompt positions during initial processing and leave later generation steps unchanged. The run specification pairs same-item sham and intervention measurements, labels observer measurements separately, and preserves the random-direction and observer conditions. From these records, we estimate finite edge-corrected probit d$'$, tied-score AUROC, and calibration using one label-independent detection report. The specification can assign the observer role to a stronger model. Deterministic tests validate construction and execution, and Section~\ref{sec:results} reports the runs scored through this path across eight models at the layer-16 residual-stream site, 3{,}072 of them. Delayed and spontaneous tracks have executed on pretrained checkpoints in the exemplar suite, as Table~\ref{tab:status} records. The release now computes the trained-classifier tier of the observer bound; on the exemplar suite it returns a chance-level margin (Section~\ref{sec:validation}), and the release does not yet preserve the Track A state for later use.

We exclude GGUF models because the intervention requires direct access to PyTorch activations, and we run the experiment on GPUs.

Under the current release limits of 32 sampled examples per benchmark and two seeds, the dose-calibrated evaluation of Qwen2.5-7B-Instruct and Mistral-7B-Instruct-v0.3 requires 64 item-seed observations per model, benchmark, condition, and scope cell. This design yields 512 runs per model-benchmark pair across four conditions and two scopes, and 3,072 runs across the six model-benchmark pairs. Each intervention or random-direction run executes five model generations plus two divergence evaluations. Sham runs omit the intervention, patch, and divergence evaluations. Observer runs include two observer generations. These are execution counts read from the run specification and not power results. The final per-cell counts follow the validation-variance simulation of Section~\ref{sec:plan}.

Simulating from the validation study's item-level variance (beta-binomial resampling of the detection margin over independent hit- and false-alarm-rate item effects, $\alpha = 0.05$) gives the expected 95 percent confidence interval width of the detection margin as a function of the per-cell item count: 2.922 $d'$ units at 8 items, 1.784 at 16, 1.235 at 32, 0.841 at 64, 0.622 at 128, and 0.455 at 256. The dose-calibrated battery's item count of 32 corresponds to an expected width of 1.235, the figure Section~\ref{sec:plan} reports. The eight-model evaluation samples items per benchmark at two seeds across four conditions and two hook scopes. This yields 216 evaluations and 3{,}072 measurement runs across four conditions and two hook scopes, of which 508 form complete intervention-sham pairs in the detection family after parse failure.

\section{Formal Definitions and Estimators}\label{app:formal}

\subsection{Measurement space and conditions}

Let $\mathcal{B}$ denote the set of source benchmarks, $\mathcal{O}$ the computational objects (residual sites, attention heads, SAE features, block outputs, each with a layer index), $\mathcal{A}$ the intervention operators with dose parameter $\delta \in \mathcal{D}$, $\mathcal{Q}$ the probe families, and $\mathcal{T} = \{A, B, B_{\mathrm{masked}}, C\}$ the temporal tracks. A cell is a tuple
\begin{equation}
  c = (m, b, o, a, q, t) \in \mathcal{M} \times \mathcal{B} \times \mathcal{O} \times \mathcal{A} \times \mathcal{Q} \times \mathcal{T},
\end{equation}
where $\mathcal{M}$ is the model set. Each realized run within a cell carries an item index $i$, a seed $r$, a hook temporal scope $h \in \{\mathrm{active}, \mathrm{lifted}\}$, and a condition
\begin{equation}
  k \in \{\mathrm{int},\ \mathrm{sham},\ \mathrm{rand},\ \mathrm{obs}\},
\end{equation}
denoting intervention, paired sham, impact-matched random perturbation, and text-only observer respectively. Sham, random, and observer runs are paired with intervention runs on identical items.

\subsection{Output divergence and impact matching}

For run $u$ with baseline next-token distributions $p^{(0)}_{u,s}$ and intervened distributions $p^{(1)}_{u,s}$ at teacher-forced positions $s = 1, \dots, S_u$, the per-run output divergence is the mean Jensen--Shannon divergence
\begin{equation}
  \bar{D}_u = \frac{1}{S_u} \sum_{s=1}^{S_u} \mathrm{JS}\!\left(p^{(0)}_{u,s} \,\|\, p^{(1)}_{u,s}\right),
  \qquad
  D^{\max}_u = \max_{s} \mathrm{JS}\!\left(p^{(0)}_{u,s} \,\|\, p^{(1)}_{u,s}\right),
\end{equation}
with $\mathrm{JS}(p \| q) = \tfrac{1}{2} \mathrm{KL}(p \| m) + \tfrac{1}{2} \mathrm{KL}(q \| m)$, $m = \tfrac{1}{2}(p + q)$. The random-direction control at site $o$ is impact-matched by calibrating its norm $\eta$ so that its induced divergence matches the treatment's target:
\begin{equation}
  \eta^\ast = \arg\min_{\eta} \left| \mathbb{E}\!\left[\bar{D}(\eta)\right] - \mathbb{E}\!\left[\bar{D}_{\mathrm{int}}\right] \right|.
\end{equation}
Matching only the norm is not enough. In $\mathbb{R}^d$ a random direction lands nearly orthogonal to the feature manifold, so at equal norm it does less downstream damage, and the control comes out weaker than the treatment it is meant to match.

\subsection{Detection: discrimination against sham}

Within a cell, let $H = \Pr(\text{report change} \mid k = \mathrm{int})$ and $F = \Pr(\text{report change} \mid k = \mathrm{sham})$ over paired items. Detection is summarized by the probit contrast and tied-score AUROC of Equation~\ref{eq:dprime}, where $g$ is a graded detection score (verbalized probability or binary report). A raw yes-rate is never the detection estimate. A model answering ``yes'' on every run attains $H = F = 1$ and $d' = 0$. We report raw rates descriptively, and detection is always the paired contrast against sham.
For finite samples, rates at 0 or 1 are replaced by $1/(2N)$ or $1-1/(2N)$, respectively, before applying the probit. Interval estimates use a nonparametric bootstrap clustered on items, resampling item pairs rather than runs, so that item difficulty does not confound discrimination.

\subsection{Margins and the reportability profile}

For probe family $q$ with score function $s_q$, define the control margins
\begin{equation}
  \mu^{\mathrm{rand}}_q = \mathbb{E}\!\left[s_q^{\mathrm{int}} - s_q^{\mathrm{rand}}\right],
  \qquad
  \mu^{\mathrm{obs}}_q = \mathbb{E}\!\left[s_q^{\mathrm{int}} - \max_{w \in \mathcal{W}} s_q^{\mathrm{obs}, w}\right],
\end{equation}
where $\mathcal{W}$ indexes the observer family: the same model zero-shot, a stronger open-weight model, and a supervised classifier trained on held-out visible outputs. The cell's reportability profile is the vector
\begin{equation}
  R_c = \left(d'_c,\ \mathrm{AUROC}_c,\ \mu^{\mathrm{rand}}_{q,c},\ \mu^{\mathrm{obs}}_{q,c}\right)_{q \in \mathcal{Q}},
\end{equation}
and no scalar composite of $R_c$ is defined.

\subsection{Dose-response model and the leakage decomposition}

Intervention strength is experimenter-set. With dose ladder $\delta \in \mathcal{D}$ as the perturbation covariate, detection follows a hierarchical logistic model
\begin{equation}
  \operatorname{logit} \Pr\!\left(y_{u} = 1\right)
  = \beta_0 + \beta_1 z_u + z_u \, g\!\left(\delta_u\right) + b_{\mathrm{item}(u)} + b_{\mathrm{site}(u)} + b_{\mathrm{model}(u)},
  \qquad u \in \mathcal{U}_{\mathrm{used}},
\end{equation}
with $z_u \in \{0, 1\}$ the intervention indicator, $\delta_u$ the experimenter-set dose, $g$ a monotone increasing spline with $g(0) = 0$, and Gaussian random effects $b$. Dose enters only in interaction with $z$, since a sham run has no intervention whose magnitude could vary. The fit is restricted to $\mathcal{U}_{\mathrm{used}}$, the used-object stratum, defined by the per-run causal-relevance patching effect of the intervened object on the task answer exceeding a fixed threshold. That effect is measured on the intervention side, and each sham run inherits the classification of the intervention run it is paired with, so the stratum is a set of pairs rather than of individual runs. Restricting to that stratum is itself a restriction on a post-treatment quantity, and we treat it as a subgroup definition rather than as a causal adjustment.

Two things follow, and both are limits rather than conveniences. First, the model is written on the dose rather than on the realized divergence $\bar{D}$, precisely because $\bar{D}$ is a consequence of treatment; the earlier form of this model conditioned on it and could not support the interpretation we wanted from it. Second, $\beta_1$ is a log-odds contrast while $m_0$ of Equation~\ref{eq:m0} is a difference of probabilities, so the two are not the same number. Zero dose is the point at which the perturbation, and hence the divergence it causes, vanishes, so the zero-dose contrast and the zero-divergence limit of Equation~\ref{eq:m0} name the same quantity; we estimate it at zero dose because dose is set by the experimenter. We define $m_0$ as the population-marginal contrast at zero dose,
\begin{equation}
  m_0 = \mathbb{E}_{b}\!\left[\operatorname{logit}^{-1}\!\left(\beta_0 + \beta_1 + b\right) - \operatorname{logit}^{-1}\!\left(\beta_0 + b\right)\right],
\end{equation}
integrating the random effects over their fitted distribution rather than evaluating at $b = 0$. Nothing in the specification forces $m_0 \leq \bar{m}$, and a leakage share outside $[0, 1]$ is possible under sampling noise or a nonmonotone relationship, so $\lambda$ is a descriptive decomposition of a behavioral margin and not an identified mechanism. Section~\ref{sec:res-dose} states why we report neither $m_0$ nor $\lambda$ for this roster.

\subsection{Variance components}

For the variance-component question, the profile components are decomposed by crossed random effects
\begin{equation}
  s_{u} = \mu + b_{\mathrm{domain}} + b_{\mathrm{object}} + b_{\mathrm{item}} + b_{\mathrm{seed}} + \varepsilon_u,
\end{equation}
with intraclass correlations
\begin{equation}
  \mathrm{ICC}_{\mathrm{domain}} = \frac{\sigma^2_{\mathrm{domain}}}{\sigma^2_{\mathrm{domain}} + \sigma^2_{\mathrm{object}} + \sigma^2_{\mathrm{item}} + \sigma^2_{\mathrm{seed}} + \sigma^2_\varepsilon},
\end{equation}
and analogously for objects. The question is whether $\mathrm{ICC}_{\mathrm{domain}}$ and $\mathrm{ICC}_{\mathrm{object}}$ exceed $\mathrm{ICC}_{\mathrm{seed}}$. We report interval estimates together with the effect sizes they bound, because an interval excluding equality establishes no meaningful difference on its own. This analysis replaces cell-wise significance tests entirely.

\subsection{Confidence scoring}

Verbalized confidence $\hat{p}_u \in [0, 1]$ is scored by the Brier score and expected calibration error over $J$ equal-mass bins,
\begin{equation}
  \mathrm{BS} = \frac{1}{n} \sum_u \left(\hat{p}_u - y_u\right)^2,
  \qquad
  \mathrm{ECE} = \sum_{j=1}^{J} \frac{n_j}{n} \left| \mathrm{acc}(j) - \mathrm{conf}(j) \right|,
\end{equation}
and by selective prediction: the AUROC of $\hat{p}$ for the model's own introspective correctness. Verbalized confidence and token-level probabilities are distinct quantities and are reported separately.

\subsection{Sensitivity anchors}

For each site, a linear probe $w^\top x + b$ is trained on held-out activations $x$ downstream of the intervention to classify intervention presence (and, where ground truth exists, identity). Its accuracy margin $m_{\mathrm{probe}}$ measures how much intervention information a linear read-out demonstrably recovers. It is a lower bound on what is linearly available, not a ceiling: a better probe or a better read-out position could do more, as the downstream re-harvest of Section~\ref{sec:res-sensitivity} in fact does. The extraction ratio
\begin{equation}
  \rho = \frac{\bar{m}}{m_{\mathrm{probe}}}
\end{equation}
locates verbal reportability against that ceiling. When $m_{\mathrm{probe}} \approx 0$ there was nothing linearly available to report, so a behavioral null is uninterpretable and the site drops out of the primary analysis. When $\rho$ approaches 1, the output channel is recovering nearly all of the linearly available intervention information. Nonlinear verbal extraction or sampling noise can place estimates of $\rho$ outside $[0, 1]$. As with $\lambda$, we report the raw estimate and interval and truncate only the headline share. The known-positive steering-aware model supplies the corresponding end-to-end instrument check.
The ratio $\rho$ is the analogue of metacognitive efficiency, the M-ratio meta-$d'/d'$, strengthened by a linear probe that supplies a cleaner capacity bound. It inherits that literature's caution about ratios with small denominators \citep{maniscalco2012signal,fleming2014how,fleming2017hmetad}.

\section{Estimands, Exclusions, and Scoring Rules}\label{app:analysis}

We define the analysis population, estimands, models, exclusion rules, and outcome-to-claim mappings using the notation of Appendix~\ref{app:formal}. For domain, layer, operator, and track analyses, we apply false-discovery control.

\subsection{Population of cells and executed tracks}

These estimands, exclusion rules, and scoring procedures govern every scored cell. The population comprises nine models evaluated at residual-stream sites in the knowledge, commonsense, and arithmetic benchmark domains. Qwen2.5-7B-Instruct and Mistral-7B-Instruct-v0.3 also use the dose ladder, per-site calibrated random-direction control, and two sensitivity anchors from Section~\ref{sec:plan}. Within each eligible model, benchmark, site, operator, probe, and track cell, we pair identical items across intervention, sham, impact-matched random-direction, and observer conditions. We cross these conditions with active-through-probe and lifted-at-probe hook scopes. Intervention dose is calibrated by site. The structured run specification fixes model revision, decoding settings, item identity, probe text, site, operator, dose, seed, condition, and hook scope.

In the notation of Appendix~\ref{app:formal}, the analysis population is the subset
\begin{equation}\label{eq:analysis-population}
  \mathcal{C}_{\mathrm{eval}}
  \subset \mathcal{M} \times \mathcal{B} \times \mathcal{O}_{\mathrm{resid}}
  \times \mathcal{A} \times \mathcal{Q} \times \mathcal{T},
\end{equation}
subject to the validation and exclusion rules below. The current release executes Track A under both hook temporal scopes and runs the masked-context reflection variant on the immediate path. This variant is not a validated delayed Track B execution. We include Tracks B and C only after validating their distinct execution paths. A release without validated Track B or Track C observations therefore cannot estimate the full ordering across Tracks A, B, and C.

\subsection{Contrasts and estimands}

Let $\Delta^{\mathrm{det}}_c$ denote the intervention-minus-sham detection margin in cell $c$, summarized by the finite-sample $d'$ and tied-score AUROC in Equation~\ref{eq:dprime}. Let $\mu^{\mathrm{obs}}_{q,c}$ and $\mu^{\mathrm{rand}}_{q,c}$ denote the observer and impact-matched random-direction margins defined in Appendix~\ref{app:formal}.

\paragraph{Sensitivity without semantic access.}
The contrast is the joint ordinal event
\begin{equation}\label{eq:contrast-sensitivity}
  \Delta^{\mathrm{det}}_c > 0
  \quad\text{and}\quad
  \mu^{\mathrm{obs}}_{\mathrm{reconstruction},c} \leq 0.
\end{equation}
Detection is evaluated against paired sham runs, and reconstruction is evaluated against the maximum over the observer family defined in Appendix~\ref{app:formal}. What is at issue is the conjunction, not either component alone.

\paragraph{Domain and object structure.}
Using the crossed variance-component model in Appendix~\ref{app:formal}, the estimands are
\begin{equation}\label{eq:contrast-structure}
  \Delta_{\mathrm{domain}}
  = \mathrm{ICC}_{\mathrm{domain}}-\mathrm{ICC}_{\mathrm{seed}},
  \qquad
  \Delta_{\mathrm{object}}
  = \mathrm{ICC}_{\mathrm{object}}-\mathrm{ICC}_{\mathrm{seed}}.
\end{equation}
Structure at the level of domain and object is established only when the interval estimates for both contrasts exclude equality on the positive side. Item-level variance remains in the crossed model and is not converted into a collection of cell-wise tests.

\paragraph{Temporal decay.}
For every pair of validated and executed tracks, define the within-cell detection-margin contrasts
\begin{equation}\label{eq:contrast-decay}
  \Delta_{AB,c}=\Delta^{\mathrm{det}}_{A,c}-\Delta^{\mathrm{det}}_{B,c},
  \qquad
  \Delta_{BC,c}=\Delta^{\mathrm{det}}_{B,c}-\Delta^{\mathrm{det}}_{C,c}.
\end{equation}
Decay across the tracks would make both contrasts positive. The masked-context condition identifies the contribution of self-reading only when paired with a validated delayed Track B path. Until we validate the execution of Tracks B and C, Equation~\ref{eq:contrast-decay} cannot estimate the full ordering, and the data do not support a temporal-decay conclusion.

\paragraph{Reportability apart from task damage.}
At a dose $\delta$ on the calibrated ladder, let
\begin{equation}\label{eq:contrast-damage}
  \begin{aligned}
    \Delta^{\mathrm{det}}_c(\delta)
    &= \Pr(y=1\mid \mathrm{int},c,\delta)-\Pr(y=1\mid \mathrm{sham},c), \\
    \Delta^{\mathrm{task}}_c(\delta)
    &= \mathbb{E}[s^{\mathrm{int}}_{\mathrm{task}}-s^{\mathrm{base}}_{\mathrm{task}}\mid c,\delta].
  \end{aligned}
\end{equation}
The question is whether the detection margin is a monotone function of task damage. Evidential weight is assigned to doses at which task accuracy is unchanged, rather than to large margins observed only when task performance is damaged.

\paragraph{Leakage bound.}
The estimands are the zero-divergence margin $m_0$ and first-order-leakage share $\lambda$ in Equation~\ref{eq:m0}. The hierarchical dose-response contrast estimates $m_0$ by extrapolation along experimenter-set dose rather than by conditioning on observed divergence. Leakage is present when $m_0<\bar{m}$ and hence $\lambda>0$; neither ordering is guaranteed by the specification, and sampling noise can place $\lambda$ outside the unit interval. Any access interpretation rests on $m_0$ in the causally used-object stratum. Section~\ref{sec:res-dose} states why neither quantity is reported for this roster.

\subsection{Analysis models and uncertainty}

Detection point estimates use the finite edge-corrected probit $d'$ and tied-score AUROC of Equation~\ref{eq:dprime}. Their intervals use a nonparametric bootstrap clustered on items: each replicate samples complete item pairs with replacement and carries the paired intervention and sham observations together. Control margins are computed within matched cells. Verbalized confidence is assessed with the Brier score and ECE over equal-mass bins as specified in Appendix~\ref{app:formal}; calibration is an instrument check rather than a claim about a model.

For every cell and condition, we define the report-parse rate as valid scorable probe responses divided by attempted probe responses. We report its numerator and denominator and use it as a manipulation check. We neither treat parse failure as incidental missingness nor impute it from other runs.

\paragraph{Human validation of reconstruction scoring.}
No reconstruction probe has been scored. Any reconstruction claim requires a human-validated subset scored by multiple raters, with the rater count, sample size, and agreement threshold fixed before that claim is made. The reconstruction rater protocol is specified but not exercised in this release: the family produced no scorable output, because these interventions carry no ground-truth concept label to score a reconstruction against.

\paragraph{Human validation of the Track C rubric.}
The condition-blinded Track C rubric is a rule-based proxy for human ratings. It has scored the exemplar suite's two completed pairs but has not been validated against human raters. We base no Track C claim on it. Validating it would require a fixed rater count, sample size, and agreement threshold set before any Track C claim is made, and Appendix~\ref{app:excluded} gives the concrete reason that validation is a precondition rather than a formality: the rubric flags the sham condition at the same rate as the intervened one, which is the signature of an instrument responding to vocabulary rather than to the intervention.

The leakage decomposition uses the hierarchical logistic dose-response model in Appendix~\ref{app:formal}, with a monotone dose spline constrained to zero at zero dose and crossed item, site, and model effects. The variance components use the crossed random-effects decomposition over domain, object, item, seed, and residual variation defined in the same appendix. Inference throughout uses hierarchical partial pooling, and the domain, layer, operator, and track analyses use false-discovery control.

\subsection{Exclusions and corruption handling}

The following exclusion and corruption rules govern the analysis.
\begin{itemize}
  \item A computational object enters the analysis only after an object-specific test establishes that the intervention changes the declared object and uses a supported operation.
  \item Malformed generation and output corruption are retained as outcomes, flagged by cell, and reported as corruption rates. They are never coded as evidence of introspective access. Discrimination is reported both with these runs retained and in the sensitivity analysis that excludes them.
  \item An unparseable probe report is excluded, together with its pair, from the complete-pair discrimination analysis that we report throughout, and is retained as zero report evidence in the inclusive analysis that we report as the robustness check. The complete-pair coding is the primary one and the inclusive coding is the sensitivity analysis, matching the usage in the Results. Both analyses are reported, together with the report-parse rate for every cell and condition. If excluding an unparseable report breaks an intervention-sham pair, the entire incomplete pair is excluded from the paired sensitivity estimate and its bootstrap interval.
  \item A site whose baseline-replacement patch indicates that the selected object is causally inert for the task is excluded from the used-object estimate of $m_0$. Its descriptive results and inert-site classification remain reportable.
  \item A site with $m_{\mathrm{probe}}\approx 0$ is excluded from behavioral-null interpretation because intervention information was not shown to be linearly available at the sensitivity anchor. The site remains in instrument-diagnostic reporting.
  \item Incomplete intervention-sham pairs do not contribute to paired discrimination or its bootstrap interval. The reason and count are reported rather than repaired by unpaired substitution.
  \item Quantized and full-precision runs are not pooled. Runs with an unrecorded model revision, decoding specification, probe text, condition, site, dose, seed, or hook scope do not enter the analysis.
\end{itemize}

\subsection{Outcome-to-claim rules}

We limit interpretation to the claim levels below. For mixed results, we report the lowest claim level whose full conditions are satisfied.
\begin{itemize}
  \item If the known-positive instrument check fails, or if $m_{\mathrm{probe}}\approx 0$ at a site, a behavioral null at that site is uninformative about reportability. It licenses only an instrument-sensitivity diagnosis.
  \item If detection does not exceed paired sham despite valid sensitivity anchors, the result bounds reportability under the tested model, cell, dose, probe, and executed track conditions. It does not license a claim that the model lacks all introspective access.
  \item If detection exceeds sham but reconstruction does not exceed the observer bound, the result licenses an interpretation of perturbation sensitivity without demonstrated semantic access.
  \item If margins occur only with task damage, measurable output divergence, or performance no better than the observer bound, the result licenses an artifact-sensitivity or first-order-leakage account, not second-order access.
  \item If detection and reconstruction exceed their controls, the known-positive and linear-probe anchors succeed, and a positive used-object $m_0$ persists near zero output divergence, the result licenses controlled evidence of reportability and behavioral evidence consistent with second-order access. It does not establish consciousness, subjective experience, moral status, or a human-like self-model.
  \item Positive domain and object contrasts license a claim that the reportability profile has structure beyond seed variation in the evaluated population. Failure of either contrast leaves that component unsupported.
  \item Positive track contrasts license temporal decay only for track pairs whose distinct execution paths were validated and executed. Immediate-path masked-context results alone do not license a Track A, Track B, Track C decay claim.
  \item A nonmonotone detection relationship with informative unchanged-accuracy doses separates reportability from task damage. A relationship confined to damaged outputs does not.
  \item A leakage share whose interval lies above zero is consistent with aggregate detection containing a first-order-leakage component. It is a decomposition of a behavioral margin, so it does not on its own identify the mechanism, and the ordering it assumes is not guaranteed by the specification. The access-relevant quantity remains the used-object estimate of $m_0$, whether or not the aggregate margin is positive.
\end{itemize}

\end{document}